\documentclass[parskip=full,12pt]{article}

\usepackage{soul}
\usepackage{amssymb}
\usepackage{natbib}
\usepackage{amsthm}
\usepackage{etoolbox}
\usepackage{lmodern}
\usepackage{amsmath,bm} 
\numberwithin{equation}{section}
\usepackage{mathtools}

\usepackage{siunitx}
\usepackage{color, soul}
\usepackage{textcomp}
\usepackage{graphicx}
\usepackage{caption}
\usepackage{indentfirst}
\usepackage{cite}
\usepackage{url}
\usepackage{titlesec}
\usepackage[top=1in, bottom=1.3in, left=0.875in, right=0.875in]{geometry}
\usepackage[linesnumbered,ruled]{algorithm2e}
\usepackage[titletoc,toc,title]{appendix}
\usepackage{tikz}
\usepackage{multicol}
\usepackage{multirow}
\usepackage{bbm}
\usepackage{booktabs}
\usepackage{hyperref}
\usepackage{xcolor}
\usepackage{listings}
\definecolor{codegreen}{rgb}{0,0.6,0}
\definecolor{codegray}{rgb}{0.5,0.5,0.5}
\definecolor{codepurple}{rgb}{0.58,0,0.82}
\definecolor{backcolour}{rgb}{0.95,0.95,0.92}

\lstdefinestyle{mystyle}{
    backgroundcolor=\color{backcolour},   
    commentstyle=\color{codegreen},
    keywordstyle=\color{magenta},
    numberstyle=\tiny\color{codegray},
    stringstyle=\color{codepurple},
    basicstyle=\ttfamily\footnotesize,
    breakatwhitespace=false,         
    breaklines=true,                 
    captionpos=b,                    
    keepspaces=true,                 
    numbers=left,                    
    numbersep=5pt,                  
    showspaces=false,                
    showstringspaces=false,
    showtabs=false,                  
    tabsize=2
}
\usetikzlibrary{er,positioning,arrows.meta,calc}
\usetikzlibrary{er,positioning,arrows.meta,arrows,shapes,calc}
\tikzstyle{block} = [rectangle, draw, fill=white, 
    text width=7cm, text centered, rounded corners, minimum height=3em]

\tikzset{
     arrow/.style = { thick,  ->, >=Triangle},
}
\usepackage[labelformat=simple]{subcaption}

\def\spacingset#1{\renewcommand{\baselinestretch}%
{#1}\small\normalsize} \spacingset{1}

\def\spacingset#1{\renewcommand{\baselinestretch}%
{#1}\small\normalsize} \spacingset{1}
\SetKwComment{Comment}{/* }{ */}

\title{Automated Machine Learning in Insurance}

\author{}
\author{Panyi Dong\thanks{Actuarial and Risk Management Sciences, University of Illinois at Urbana-Champaign, 1409 W. Green Street (MC-382), Urbana, IL, 61801, USA. Email: \texttt{panyid2@illinois.edu}.}\and Zhiyu Quan\thanks{Actuarial and Risk Management Sciences, University of Illinois at Urbana-Champaign, 1409 W. Green Street (MC-382), Urbana, IL, 61801, USA. Email: \texttt{zquan@illinois.edu}.}}
    
\begin{document}

\setstcolor{red}

\maketitle
	
\begin{abstract}
		
Machine Learning (ML) has gained popularity in actuarial research and insurance industrial applications. However, the performance of most ML tasks heavily depends on data preprocessing, model selection, and hyperparameter optimization, which are considered to be intensive in terms of domain knowledge, experience, and manual labor. Automated Machine Learning (AutoML) aims to automatically complete the full life-cycle of ML tasks and provides state-of-the-art ML models without human intervention or supervision. This paper introduces an AutoML workflow that allows users without domain knowledge or prior experience to achieve robust and effortless ML deployment by writing only a few lines of code. This proposed AutoML is specifically tailored for the insurance application, with features like the balancing step in data preprocessing, ensemble pipelines, and customized loss functions. These features are designed to address the unique challenges of the insurance domain, including the imbalanced nature of common insurance datasets. The full code and documentation are available on the GitHub repository.\footnote{ \href{https://github.com/PanyiDong/InsurAutoML}{https://github.com/PanyiDong/InsurAutoML}.}
		
\vspace{0.75cm}


\end{abstract}

\newpage

\section{Introduction}\label{sec:intro}

Machine Learning (ML), as described by \citet{Mitchell1990}, is a multidisciplinary sub-field of Artificial Intelligence (AI) focused on developing and implementing algorithms and statistical models that enable computer systems to perform data-driven tasks or make predictions through ``leveraging data" and iterative learning processes. This data-driven approach guides the design of ML algorithms, allowing them to grasp the distributions and structures within datasets and unveil correlations that elude traditional mathematical and statistical methods. 
Professionals in data-related fields, such as data scientists and ML engineers, can engage in autonomous decision-making based on data and benefit from cutting-edge predictions generated by modern ML models. 


In recent decades, ML has significantly reshaped various industries and gained widespread popularity in academia due to its exceptional predictive capabilities. As summarized by \citet{jordan2015machine}, ML has made significant contributions in various fields, including robotics, autonomous driving, language processing, and computer vision. The medical and healthcare industry, as suggested by \citet{kononenko2001machine} and \citet{qayyum2020secure}, is increasingly adopting ML for applications such as medical image analysis and clinical treatments. Furthermore, ML models have significantly improved personalization and targeting, marketing strategy, and customer engagement in the marketing sector, as summarized by \citet{maMachineLearningAI2020}. \citet{guerraMachineLearningApplied2021} present the ML innovations in the banking sector, particularly in the analysis of liquidity risks, bank risks, and credit risks. Additionally, there is a growing trend in adopting ML models in the insurance sector and among actuarial researchers and industry practitioners, as evidenced by recent literature. Some recent literature advances emerging data-driven research topics in areas such as climate risks, health and long-term care, and telematics.
For instance, by combining dynamic weather information with deep learning techniques, \citet{shiLeveragingWeatherDynamics2024b} develop an enhanced predictive model, leading to improved insurance claim management. 
\citet{hartmanPredictingHighCostHealth2020} and \citet{cummingsUsingMachineLearning2022} explore various ML models on health and long-term care insurance. 
\citet{maselloUsingContextualData2023} and \citet{peirisIntegrationTraditionalTelematics2024} find that the integration of telematics through ML models can better comprehend risk characteristics. 
Other researchers focus on developing enhanced ML models to further improve predictive capabilities or address specific challenges within the insurance sector. 
\citet{charpentierReinforcementLearningEconomics2023} propose a reinforcement learning technique and explore its application in the financial sector. 
\citet{guojunCompositionalDataRegression2024} explore the use of exponential family principal component analysis in analyzing insurance compositional data to improve predictive accuracy. 
\citet{turcotteGAMLSSLongitudinalMultivariate2024} develop a generalized additive model for location, scale, and shape (GAMLSSs) to better understand longitudinal data. 
To address the excessive zeros and heavily right-skewed distribution in claim management, \citet{soEnhancedGradientBoosting} compare various ML models and propose a zero-inflated boosted tree model. 
By combining classification and regression, \citet{quanHybridTreebasedMethods2023} propose a hybrid tree-based ML algorithm to model claim frequency and severity.
In addition to advancements in ML models, researchers, such as \citet{frees2016multivariate} and \citet{charpentier2014computational}, have collected various insurance datasets. For further studies, see \citet{noll2020case}, \citet{quan2018predictive} 
and \citet{siAutomobileInsuranceClaim2022}.

In an ideal scenario, ML models are crafted to automate data-driven tasks, aiming to minimize manual labor and human intervention. However, due to their dependence on data, there is no one-size-fits-all solution, necessitating the training of a specific ML model for each task. This complexity is compounded by the vast array of ML models available, each with numerous hyperparameters controlling the learning process and impacting model performance. It even leads to the emergence of a new research field, Hyperparameter Optimization \citep{yang2020hyperparameter}, which has become increasingly vital in both academic and industrial practice. Consequently, finding the optimal hyperparameter setting for the ML model becomes a labor-intensive process reliant on extensive experience in ML. Additionally, the proliferation of digitalization has led to explosive growth in data volume and variety (number of features and observations), coupled with a decline in overall data quality. As a result, real-world datasets, especially in industrial settings, demand meticulous data preprocessing to achieve optimal performance. This data preprocessing often involves manual optimization, model-specific, and trial-and-error, further complicating the training of the ML model. These factors contribute to the difficulty of building practical ML models for individuals lacking prior experience. Even for ML experts, achieving optimal performance on new datasets can be daunting, as evidenced in Kaggle\footnote{\href{https://www.kaggle.com/competitions}{https://www.kaggle.com/competitions}} competitions. 

In this paper, we endeavor to make ML more accessible to inexperienced users and develop an inclusive learning tool that assists insurance practitioners and actuarial researchers in utilizing state-of-the-art ML tools in their daily operations. In fact, this research stems from the IRisk Lab\footnote{See IRisk Lab currently serves as an academic-industry collaboration hub, facilitates the integration of discovery-based learning experiences for students, and showcases state-of-the-art research in all areas of Risk Analysis and Advanced Analytics. Retrieved from \url{https://asrm.illinois.edu/illinois-risk-lab}} project, which facilitates the learning of ML for actuarial science students and establishes a comprehensive pipeline for automating data-driven tasks. Automated Machine Learning (AutoML), as summarized in \citet{zoller2021benchmark}, offers a solution aimed at diminishing repetitive manual tuning efforts, thus expediting ML training. Its objective is to encourage the adoption of ML across various domains, particularly among inexperienced users, by facilitating full ML life cycles and enhancing comprehension of domain datasets. Successful implementations of AutoML span across academic open-source libraries, \citet{H2OAutoML20}, \citet{feurerAutosklearnHandsfreeAutoML2022}, and industry-commercialized products, including startups like DataRobot\footnote{\href{https://www.datarobot.com/}{https://www.datarobot.com/}} and cloud services such as AWS SageMaker\footnote{\href{https://aws.amazon.com/sagemaker//}{https://aws.amazon.com/sagemaker/}}. However, existing open-source AutoML implementations may not be suitable for the insurance domain due to specific challenges, such as imbalanced datasets, a high prevalence of missing values, and scalability issues.

Our proposed AutoML pipeline tailored for the insurance domain encompasses fully functional components for data preprocessing, model selection, and hyperparameter optimization. We envision several use cases for our proposed AutoML. Firstly, it can serve as a performance benchmark for evaluating future ML creations among researchers, actuaries, and data scientists in the insurance sector. For instance, users can input datasets into our AutoML at the onset of a data project. Meanwhile, users can manually analyze the datasets, plan their subsequent steps, and produce preliminary results. Upon obtaining a prototype of the initial model, users can then compare it with the results generated by our AutoML to gauge if their model surpasses the benchmark set by our AutoML. This utilization of AutoML as a benchmark aids in standardizing and advancing ML methodologies within the dynamic landscape of the insurance industry, including academic research. Secondly, our proposed AutoML generates training history as a byproduct, offering insights that users may have overlooked. For example, users can review the training process and the results recorded by AutoML over time. This feature can provide insights that align with user experiments or intuition, or sometimes it can present counterintuitive findings that prompt users to reconsider their approach. Thirdly, our AutoML, designed with flexibility in the search space and optimization algorithms, can seamlessly incorporate future innovations while maintaining its strength in automation. For experienced users seeking to unlock the full potential of our AutoML, its flexible design enables them to leverage tuning results to gain a deeper understanding of the underlying data structure. Such insights can then serve as a guideline for manually reducing the search complexity, ultimately facilitating the attainment of optimal performance in time and computation. Finally, our research has real-world applicability and is available as an open-source tool, making it free for users to implement in practical scenarios. For instance, in the life-cycle of insurance business operations, our AutoML can offer tool sets to insurers to improve underwriting processes, optimize pricing strategies, enhance risk management practices, and boost operational efficiency, cost reduction, and customer satisfaction. Thus, AutoML presents a promising opportunity for insurance companies to leverage advanced ML models and unlock the full potential of their data. Additionally, we believe that our open-source AutoML can serve as an educational tool for university students and a benchmark-building resource for academic research.

The paper is structured in the following sections: Section \ref{sec:workflow} provides an overview of the general AutoML workflow and formulates the processes involved in model selection and hyperparameter optimization. Section \ref{sec:InsurAutoML} focuses specifically on our AutoML design tailored for the insurance domain, emphasizing the integration of sampling techniques and ensemble learning strategies to address the unique issues in insurance data. To showcase the feasibility of our AutoML in the insurance domain, Section \ref{sec:exp} presents experiment results demonstrating the performance of our AutoML on various insurance datasets and compares it with existing research. Notations utilized throughout the paper can be found in the Appendix \ref{appendix_sec:notation}. The experiments demonstrate that our AutoML carries the potential to achieve superior performance without extensive human intervention, thereby freeing practitioners and researchers from tedious manual tuning work. Section \ref{sec:conclusion} provides concluding remarks, summarizing the key findings and insights presented in the paper.

\section{The Concept of AutoML}\label{sec:workflow}

Most ML algorithm architectures characterize an ML model $M$ by a set of parameters $\theta$ and hyperparameters $\lambda$. Parameters $\theta$ are essential components of the ML model structure, representing values that are trainable during model calibration and estimated from data, such as the estimated prediction values of the leaf nodes in tree-based models or the weights and biases in a neural network's linear layer. In contrast, hyperparameters $\lambda$ control the model's flexibility and learning process, such as the maximum depth in tree-based models or the learning rate in neural networks. Unlike parameters $\theta$, which are determined during the training process, hyperparameters $\lambda$ are set before training and are chosen by users based on their experience and preferences. Extensive empirical experiments have shown that the careful selection of hyperparameters can significantly impact model performance. However, there exists no universal set of hyperparameters that guarantees optimal performance across all datasets, nor are there established theoretical foundations providing precise guidance for their selection.  

With a set of fixed hyperparameters $\lambda$, the optimization of the ML model $M$ utilizing dataset $\mathcal{D}=(\textbf{X}, \textbf{y})$ can be expressed as
$$
\underset{\theta}{\mathrm{argmin}}\mathcal{L}(M_{\lambda}^{\theta}(\textbf{X}), \textbf{y})
$$
where $\mathcal{L}$ denotes the loss function, which takes predictions and true values as inputs and returns a numerical value indicating the goodness-of-fit of the model. Thus, for any given loss functions $\mathcal{L}$, the model $M$ and hyperparameter $\lambda$ are critical components that control the task performance. 

In a broader sense, selecting the loss function $\mathcal{L}$ can also be considered as a hyperparameter, depending on the task's objective. Empirical experiments have demonstrated that ML models optimized for different loss functions often exhibit varying performance when evaluated with other loss functions. The loss function $\mathcal{L}$ inherently embeds the intuitions of business decision-making, making its selection crucial for effective ML deployments. For example, Mean Square Error (MSE) or Mean Absolute Error (MAE) might be more suitable for tasks such as personalized insurance loss modeling or individual fair pricing, while metrics like Percentage Error (PE) may better reflect the portfolio-level optimization. Therefore, $\mathcal{L}$ can be considered as a hyperparameter based on the specific objective of ML applications. In our AutoML implementation, we offer users the option to choose from existing loss functions or define their own based on the task requirements. We discuss this in more detail in Section \ref{subsec:lossfunction}




Model Selection and Hyperparameter Optimization (HPO) are two of the most crucial components of AutoML, both fundamentally characterized as optimization problems. The objectives are to identify the best-performing model architecture and the optimal hyperparameter settings. We divide the dataset as $\mathcal{D}=(\mathcal{D}_{train}, \mathcal{D}_{valid})$, where $\mathcal{D}_{train}=(\textbf{X}_{train}, \textbf{y}_{train})$ is the training set and $\mathcal{D}_{valid}=(\textbf{X}_{valid}, \textbf{y}_{valid})$ is the validation set. The model $M$ belongs to the model space $\mathcal{M}$, which is the collection of all appropriate models, i.e., $M\in\mathcal{M}$. In addition, we denote the default hyperparameter settings (usually provided by the built-in code) of the model architecture as $\lambda_{0}$. The naive model selection problem can then be formulated as
$$
M^{*}=\underset{M\in\mathcal{M}}{\mathrm{argmin}}\, \mathbb{E}_{\mathcal{D}\sim(\mathcal{D}_{train}, \mathcal{D}_{valid})}\mathcal{V}(\mathcal{L}, M_{\lambda_{0}}^{\theta}, \mathcal{D})
$$
where $\mathcal{V}$ denotes an evaluation process that takes a loss function $\mathcal{L}$, a initialized model $M_{\lambda_{0}}^{\theta}$ and a dataset $\mathcal{D}$. The evaluation process $\mathcal{V}$ first optimizes the model on the training set in the form of
$$
\theta^{*}=\underset{\theta}{\mathrm{argmin}}\mathcal{L}(M_{\lambda_{0}}^{\theta}(\textbf{X}_{train}), \textbf{y}_{train})
$$
and returns the evaluation loss through $L^{eval}=\mathcal{L}(M_{\lambda_{0}}^{\theta^{*}}(\textbf{X}_{valid}), \textbf{y}_{valid})$ with the optimally trained parameterized model. Thus, the process of naive model selection determines the best model architecture $M^{*}$ in the model space $\mathcal{M}$ that achieves the minimal evaluation loss $L^{eval}$. It is worth noting that naive model selection does not involve HPO which requires finding the optimal hyperparameters for all the candidate models in the model space $\mathcal{M}$. In line with the established convention of minimizing loss in optimization problems, our approach to model selection also follows the direction of minimization. Thus, for loss functions (performance metrics), such as accuracy score, $R^{2}$ score, and Area Under the Curve (AUC) score that aim to maximize values, we use negative values to maintain coherence in the optimization direction.


For a fixed model $M$, if the hyperparameter set $\lambda$ consists of $N$ tunable hyperparameter, an $N$ dimensional hyperparameter space $\bm{\Lambda}^{M}=\Lambda_{1}^{M}\times\Lambda_{2}^{M}\times\ldots\times\Lambda_{N}^{M}$ can be configured, where $\Lambda_{j}^{M}$, $j=1, \dots, N$ represents the available range of hyperparameter $j$. A specific hyperparameter set $\lambda^{M}\in\bm{\Lambda^{M}}$ is a $N$ dimension vector within the hyperparameter space $\bm{\Lambda}^{M}$. HPO aims to find the optimal hyperparameter set $\lambda^{M*}$ such that
$$
\lambda^{M*}=\underset{\lambda^{M}\in\bm{\Lambda}^{M}}{\mathrm{argmin}}\, \mathbb{E}_{\mathcal{D}\sim(\mathcal{D}_{train}, \mathcal{D}_{valid})}\mathcal{V}(\mathcal{L}, M_{\bm{\lambda}^{M}}^{\theta}, \mathcal{D})
$$
where $\mathcal{V}$ and $\mathcal{L}$ refer to the same evaluation process and the loss function defined in the naive model selection process.

As suggested by \citet{HE2021106622}, many solutions for Model Selection and HPO, especially in the field of Neural Networks (NNs) where the Model Selection is transformed into the Neural Architecture Search (NAS) problem, rely on a two-stage framework that separates the model architecture search and hyperparameter optimization. The two-stage optimization can be expressed as:
\begin{align*}
    M^{*}&=\underset{M\in\mathcal{M}}{\mathrm{argmin}}\, \mathbb{E}_{\mathcal{D}\sim(\mathcal{D}_{train}, \mathcal{D}_{valid})}\mathcal{V}(\mathcal{L}, M^{\theta}_{\lambda_{0}}, \mathcal{D})\\
    \lambda^{M*}&=\underset{\lambda^{M}\in\bm{\Lambda}^{M}}{\mathrm{argmin}}\, \mathbb{E}_{\mathcal{D}\sim(\mathcal{D}_{train}, \mathcal{D}_{valid})}\mathcal{V}(\mathcal{L}, {M^{*}}_{\bm{\lambda}^{M}}^{\theta}, \mathcal{D})
\end{align*}

In the first stage, the model $M^{*}$ is found using default hyperparameter settings $\lambda_{0}$. In the second stage, the HPO is performed with the restricted model $M^{*}$ provided by the first stage to find the optimal hyperparameter set $\lambda^{M*}$. The two-stage optimization generates an optimized ML model $M_{\bm{\lambda}^{M*}}^{*}$. While this strategy is efficient from a time and computation perspective, it may lead to local optima in both stages rather than the global optimum.

Considering that the insurance industry prioritizes accuracy over computational efficiency, we adopt a similar approach to Auto-WEKA, proposed by \citet{Thornton2013}. This approach merges the problems of Model Selection and HPO into a \textit{combined algorithm selection and hyperparameter optimization} (CASH) problem. The CASH framework treats the models as tunable hyperparameters, generating a hierarchical nested hyperparameter space. Therefore, the optimization of the CASH problem can be formulated as
\begin{align*}
    M^{*}_{\lambda^{*}}&=\underset{M\in\mathcal{M}, \lambda^{M}\in\bm{\Lambda}^{M}}{\mathrm{argmin}}\, \mathbb{E}_{\mathcal{D}\sim(\mathcal{D}_{train}, \mathcal{D}_{valid})}\mathcal{V}(\mathcal{L}, M^{\theta}_{\lambda^{M}}, \mathcal{D})\\
    &=\underset{(M,\lambda^{M})\in\mathcal{C}^{M}}{\mathrm{argmin}}\, \mathbb{E}_{\mathcal{D}\sim(\mathcal{D}_{train}, \mathcal{D}_{valid})}\mathcal{V}(\mathcal{L}, M^{\theta}_{\lambda^{M}}, \mathcal{D})
\end{align*}
where $\mathcal{C}^{M}=\mathcal{M}\times\bm{\Lambda}^{M}$ denotes a conjunction space of model space $\mathcal{M}$ and hyperparameter space $\bm{\Lambda}^{M}$, and $(M,\lambda^{M})$ is a sample in the conjunction space. By combining Model Selection and HPO, the CASH framework is able to find the global optimum, albeit at the expense of increased computational resources.

\section{Insurance Domain-Specific AutoML}\label{sec:InsurAutoML}

When deploying ML models in real-world scenarios, practitioners and researchers recognize that the unique and complex structure of the insurance industry can impede ML innovations. One of the most pressing challenges for ML in the insurance sector is data quality, which is affected by several factors, including legacy systems, operation-oriented data collection practices, and inadequate database management. Firstly, many insurance companies rely on outdated legacy systems that lack the capability to store and process modern data formats, resulting in inconsistent, incomplete, or inaccurate data. Integrating data from these systems with newer technologies can be complex and error-prone. Secondly, traditional data collection practices in the insurance industry are operation-oriented rather than analysis-oriented. Data is often gathered to support specific operational processes without considering its potential use for analytical or predictive modeling. This leads to fragmented, siloed, or insufficiently detailed data, detrimental to effective ML applications. Thirdly, insufficient database management practices further degrade data quality. Issues such as duplicate records, missing values, and outdated information are common, hindering the performance and reliability of ML models. The performance of ML models is heavily dependent on data quality. High-quality data facilitates easier and more efficient model building and results in higher inference accuracy. However, data quality is primarily determined by collection procedures, which are beyond the control of ML models. In response to these challenges, our AutoML incorporates comprehensive data preprocessing techniques to extract as much valuable information as possible through automated trial and error, which is otherwise infeasible without extensive manual work. Our AutoML seeks to build models that are robust, accurate, and capable of delivering reliable predictions despite the inherent challenges associated with data quality. This systematic approach allows us to maximize the value extracted from the available data, ultimately leading to better insights and decision-making in the insurance sector.

Another unique challenge in the insurance sector is the problem of imbalanced data. When modeling future claims, for example, claim events are relatively rare, with most policyholders not experiencing any claims. Typically, the claim events constitute less than 10\% of all policies, and in some cases, such as policies covering catastrophic events, the proportion can be as low as 0.1\%. This phenomenon is referred to as an imbalanced data problem in ML, where a single class comprises the majority of all observations. Specifically, in the field of imbalance learning, the class taking the dominant proportion is typically referred to as \textit{majority class} while others denote \textit{minority class}. In certain domains, observations within the minority class or rare observations are referred to as \textit{outliers} \citep{hodge2004survey}. These outliers are often treated as noise to be removed or merged into other classes. However, in the insurance industry, these minority class or rare observations corresponding to the claim events contribute as a crucial estimation of financial liabilities or \textit{pure premiums} in the actuarial term. Consequently, accurately estimating the minority class or rare observations is crucial for the insurance domain. However, the majority of ML models, by their designs, assume equal contributions from each observation, regardless of whether they belong to the majority or minority classes. As a result, many ML models, without appropriate modifications, underperform on imbalanced datasets. The problem of imbalanced data has caught the attention of researchers across diverse fields, leading to the development of various solutions summarized by \citet{He2009} and \citet{Haixiang2017}. These imbalance learning techniques, in general, can be categorized as: (1) Sampling methods; (2) Cost-sensitive methods; (3) Ensemble methods; and (4) Kernel-based methods and Active Learning methods. Our AutoML incorporates a series of sampling methods as a critical data preprocessing component to balance majority and minority classes. In addition, advancements in imbalance learning within actuarial science, such as cost-sensitive loss functions \citep{huImbalancedLearningInsurance2022a, soCOSTSENSITIVEMULTICLASSADABOOST2021a} can be integrated into our loss function. Furthermore, as summarized by \citet{Sagi2018},  ensemble learning not only achieves state-of-the-art performance, but also has the potential to address the imbalance problem effectively by combining multiple ML models into a predictive ensemble model. The multiple evaluation pipelines fitted during our AutoML training can naturally serve as candidates to form an ensemble.

Our AutoML focuses on structured data, such as tabular data from databases or CSV files, and is particularly designed for supervised learning problems, specifically regression and classification. While many other types of AutoML handle unstructured data, like audio, text, video, and images, and incorporate unsupervised learning or active learning, these may lie beyond the scope of our AutoML. Instead, we aim to tailor AutoML solutions to the insurance domain to enhance the performance and adaptability of ML solutions. In the following, we introduce our AutoML from a microscopic to a macroscopic perspective. Subsection \ref{subsec:pipeline} outlines each component of the model pipeline, with a specific focus on sampling techniques. Subsection \ref{subsec:opt-prac} details how these components are interconnected and optimized within our AutoML workflow. Subsection \ref{subsec:ensemble} summarizes the integration of ensemble model as an alternative solution for imbalance learning and as the highest layer of our AutoML production. Lastly, in Subsection \ref{subsec:lossfunction}, discusses the selection of loss functions and their strategic implications in business contexts.

\subsection{Model Pipeline}\label{subsec:pipeline}

Applying the model $M$ directly to the raw dataset $\mathcal{D}$, as formulated in Section \ref{sec:workflow}, is often impractical or even impossible for several reasons. These include the diverse and complex structures of datasets, the infeasible computational run-time required for optimization, and sub-optimal performance metrics during the evaluation phase. For example, missing values can hinder the application of linear models without proper imputation methods, and the inclusion of wide-ranging features can substantially slow down the convergence rate of gradient-based optimization methods. To address these challenges, a category of techniques known as \textit{Data Preprocessing}, is employed to prepare the raw dataset before model optimization. 

Data preprocessing techniques, such as imputation and feature selection, lack dedicated evaluation metrics aside from the final model prediction metrics. Moreover, there is no universal preprocessing technique that consistently demonstrates superior performance across all datasets. As a result, these data preprocessing techniques are often data-dependent, requiring extensive manual adjustments and domain expertise. This reliance adds another layer of complexity to the construction of the optimal pipeline for any ML task, in addition to model selection and hyperparameter optimization.

In our AutoML, we incorporate five classes of data preprocessing techniques commonly employed in ML, aiming to cover a comprehensive range of data preprocessing needs. According to \citet{garcia2016big}, these data preprocessing techniques are both essential and beneficial to the ML models. In the order of pipeline fitting, we include \textit{Data Encoding}, \textit{Data Imputation}, \textit{Data Balancing}, \textit{Data Scaling}, and \textit{Feature Selection}, which we implement in our AutoML based on the reference outlined as follows:

\textbf{Data Encoding}: Data Encoding converts string-based categorical features into numerical ones, either in the format of ordinal or binary one-hot encoding. Since most ML models do not support string variables, this unified numerical representation of features ensures consistency across the dataset and prevents issues like unseen categorical variables and type inconsistencies.

\textbf{Data Imputation}: Missing values, whether due to intrinsic information loss or database mismanagement, pose significant challenges for ML. Various imputation techniques have been proposed to address the missing value issue, such as statistical methods like multiple imputation \citep{Azur2011}, non-parametric approach \citep{Stekhoven2012}, and generative adversarial networks \citep{Yoon2018}. These imputation solutions determine the optimal estimates for the missing cells and populate them accordingly. Unlike the common practice of discarding observations containing missing values, i.e., complete data analysis, which maintains the accuracy of the remaining data, imputation techniques leverage all available information, potentially enhancing the understanding of the dataset while also carrying the risk of introducing misleading imputed values. To address the uncertainty in imputation, we offer users several highly cited methods to help find the best imputation solution.

\textbf{Data Balancing}: Data balancing addresses class imbalance using sampling methods, which are especially beneficial in fields like insurance, where infrequent but significant events, such as frauds and claims, must be accurately modeled. As summarized by \citet{Batista2004}, sampling methods allow ML models to learn from these rare events better by adjusting class distributions and improving decision boundaries. 

Summarized by \citet{Chawla2002}, sampling methods are generally classified into over-sampling and down-sampling. Over-sampling increases the number of minority class instances by generating synthetic data, while under-sampling reduces the number of majority class observations.

For the dataset $\mathcal{D}=(\textbf{X}, \textbf{y})$, an imbalance problem can be described by splitting it into a majority subset $\mathcal{D}_{major}=(\textbf{X}_{major}, \textbf{y}_{major})$ and a minority subset $\mathcal{D}_{minor}=(\textbf{X}_{minor}, \textbf{y}_{minor})$, where $y=A$ for every $y\in\textbf{y}_{major}$ and $y\neq A$ for every $y\in\textbf{y}_{minor}$, such that $\mathcal{D}=\begin{bmatrix}
    \mathcal{D}_{major}\\
    \mathcal{D}_{minor}
\end{bmatrix}=(\begin{bmatrix}
    \textbf{X}_{major}\\
    \textbf{X}_{minor}
\end{bmatrix}, \begin{bmatrix}
    \textbf{y}_{major}\\
    \textbf{y}_{minor}
\end{bmatrix})$ and an imbalance ratio
$$
\dfrac{|\mathcal{D}_{major}|}{|\mathcal{D}_{minor}|}\gg 1
$$
Here, $A$ refers to the response variable defining the majority class, and $|\mathcal{D}_*|$ denotes the cardinality of dataset $\mathcal{D}_*$, indicating the number of observations. In insurance, for instance, non-fraudulent observations usually form the majority class, suggesting $A=0$. Given the imbalance between $\mathcal{D}_{major}$ and $\mathcal{D}_{minor}$, conventional ML models are usually more influenced by observations in $\mathcal{D}_{major}$ than those in $\mathcal{D}_{minor}$, resulting in inferior ML performance.

Over-sampling retains the structure of $\mathcal{D}_{major}$ and construct a sampled minority subset $\mathcal{D}_{minor}^{Y}$ such that $\mathcal{D}_{minor}\subset\mathcal{D}_{minor}^{Y}$ and
$$
\dfrac{|\mathcal{D}_{major}|}{|\mathcal{D}_{minor}^{Y}|}=R
$$
where $R$ is a pre-defined threshold or can be a hyperparameter, measuring the imbalance ratio, typically $R \approx 1$. In our AutoML, datasets with an imbalance ratio greater than $R$ are considered imbalanced, and sampling methods are applied to adjust the imbalance ratio. The additional observations in $\mathcal{D}_{minor}^{Y}\setminus\mathcal{D}_{minor}$ are synthetically generated samples that simulate the statistical properties of the observations in $\mathcal{D}_{minor}$. Common generation methods include duplication, as described by \citet{Batista2004} in the Simple Random Over-Sampling method, and linear interpolation in the feature space, as used in Synthetic Minority Over-Sampling Techniques (SMOTE) by \citet{Chawla2002}. These synthetic minority samples increase the size of the minority subset, making the majority and minority classes comparable in size and mitigating the imbalance problem caused by the disparity between the majority and minority classes.

Under-sampling, on the contrary, maintains $\mathcal{D}_{minor}$ and construct a reduced majority subset $\mathcal{D}_{majority}^{Y}$ such that $\mathcal{D}_{majority}^{Y}\subset\mathcal{D}_{majority}$ and
$$
\dfrac{|\mathcal{D}_{major}^{Y}|}{|\mathcal{D}_{minor}|}=R
$$

The subset $\mathcal{D}_{major}^{Y}$ is constructed by removing observations from $\mathcal{D}_{major}$ while preserving the statistical similarities. Most under-sampling techniques use k Nearest Neighbors (kNN) as the base learner. Tomek Link, proposed by \citet{TOMEK1976}, utilizes kNN to find the adjoining majority-minority pairs for removal. Edited Nearest Neighbors (ENN) \citep{Wilson1972} employs the predictions of kNN as majority votes to determine which majority class observations should be removed. Condensed Nearest Neighbors (CNN) by \citet{Hart1968} uses kNN as the benchmark to determine the necessary majority class observations required to generate the subset. Assuming that the majority and minority classes can be viewed as a binary classification problem and that distinctions between them can be discerned through their spatial distributions in feature space, the kNN leverages nearest neighbors as a criterion for statistical significance to identify sample importance. By removing observations that disagree with kNN predictions, majority observations that distort the smoothness of decision boundaries can be precisely eliminated, leading to smooth decision boundaries and a proper balance between majority and minority classes.


\textbf{Data Scaling}: Data scaling might not always significantly impact model performance,  but it usually accelerates convergence, especially in the case of gradient-based optimizations on skewed features. Considering the intensive computation involved in AutoML optimization, time efficiency is crucial, making the scaling of features as important as other data preprocessing components. Furthermore, some of the scaling techniques help remove outliers in features, which are beneficial to the ML models in certain scenarios. In our AutoML, we incorporate a series of common scaling techniques such as standardization and normalization are incorporated.

\textbf{Feature Selection}: In real-world applications, unprocessed features can suffer from redundancy or ambiguity, negatively affecting run-time and potentially undermining the model performance. Feature selection addresses this issue by reducing dimensions and effectively identifying a subset of valuable features. As summarized by \citet{Chandrashekar2014}, feature selection techniques can be either model-dependent (i.e., wrapper methods) or model-free (i.e., filter methods), and the effectiveness of the selection heavily relies on the datasets.

Preprocessing techniques, combined with ML models, comprise a pipeline that represents a real-world workflow of modeling tasks. To differentiate the types of hyperparameters, we extend the notations from Section \ref{sec:workflow} as follows: encoding algorithm $E$ with hyperparameter $\lambda_{E}$; imputation algorithm $I$ with hyperparameter $\lambda_{I}$; balancing algorithm $B$ with hyperparameter $\lambda_{B}$; scaling algorithm $S$ with hyperparameter $\lambda_{S}$; feature selection algorithm $F$ with hyperparameter $\lambda_{F}$; and ML model $M$ with hyperparameter $\lambda_{M}$. In our AutoML, the pipeline $\mathcal{P}$ can be initialized by the input of algorithm-hyperparameter pairs, denoted as
$$
\mathcal{P}_{0}=M_{\lambda_{M}}\circ F_{\lambda_{F}}\circ S_{\lambda_{S}}\circ B_{\lambda_{B}}\circ I_{\lambda_{I}}\circ E_{\lambda_{E}}
$$
where the initialized pipeline $\mathcal{P}_{0}$ is controlled by trainable parameters $\theta$. The pipeline can then be trained to optimize $\theta$ as follows:
$$
\theta^{*}=\underset{\theta}{\mathrm{argmin}}\mathcal{L}(\mathcal{P}_{0}(\textbf{X}), \textbf{y})
$$
where $\mathcal{D}=(\textbf{X}, \textbf{y})$ is a dataset and $\mathcal{L}$ refers to the loss function. The preprocessing techniques are applied to the datasets sequentially in the specified order, as demonstrated in the parameterization of the pipeline. This ordering demonstrated in our AutoML is a widely accepted modeling process that embeds some of the typical incentives in the domain of data science. For example, the encoding and imputation process solve the fundamental incompatibility issues that are not resolvable in subsequent operations. Further, it can be evident from the pipeline fitting that, while the pipeline encompasses the entire life-cycle of modeling tasks, the selection of algorithms and their corresponding hyperparameters still relies on manual decisions or automated optimization. The automated optimization of such algorithm-hyperparameter pairs underscores the term ``Auto" in AutoML.

\subsection{Automated Optimization}\label{subsec:opt-prac}

To achieve automated optimization, we adopt the framework of CASH, as formulated in Section \ref{sec:workflow}, and extend it to the preprocessing-modeling space. Given the preprocessing-modeling pipeline summarized in Subsection \ref{subsec:pipeline}, we construct the conjunction space for Data Encoding, Data Imputation, Data Balancing, Data Scaling, Feature Selection, denoted as $\mathcal{C}^{E}$, $\mathcal{C}^{I}$, $\mathcal{C}^{B}$, $\mathcal{C}^{S}$, $\mathcal{C}^{F}$, respectively. In addition, the optimal algorithm-hyperparameter pairs can be identified by exploring the documentation from implemented libraries or packages. This allows us to incorporate the pipeline fitting into the objective function $\mathcal{V}$, extending it as follows:
$$
\mathcal{V}(\mathcal{L}, M_{\lambda_{M}}\circ F_{\lambda_{F}}\circ S_{\lambda_{S}}\circ B_{\lambda_{B}}\circ I_{\lambda_{I}}\circ E_{\lambda_{E}}, \mathcal{D})
$$

\begin{figure}[htbp]
\centering
\includegraphics[width=0.9\linewidth]{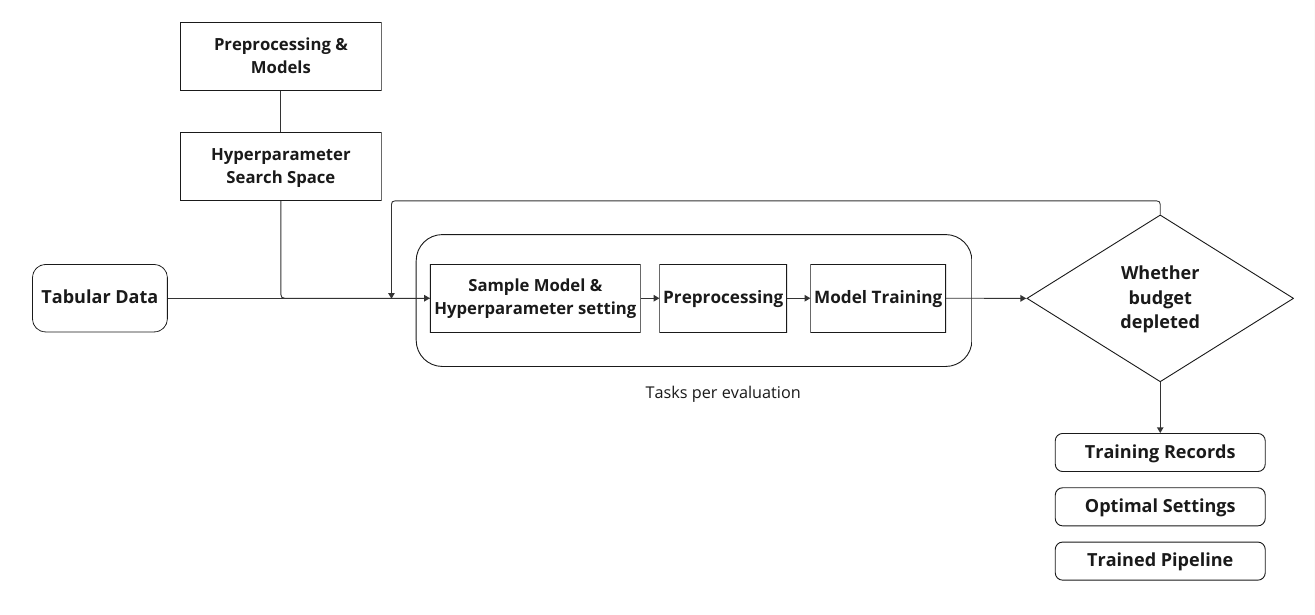}
\caption{An illustration of AutoML workflow}
\label{fig::workflow}
\end{figure}

In the following, we demonstrate the optimization strategy adopted in our AutoML, focusing on supervised learning tasks on tabular datasets. Figure \ref{fig::workflow} illustrates the automated optimization workflow, where a space of preprocessing techniques and ML models, along with the corresponding hyperparameter space for each method, is constructed and stored prior to optimization. This space is usually denoted as the \textit{Search Space} and can be written as
$$
\mathcal{U}=\mathcal{C}^{E}\times\mathcal{C}^{I}\times\mathcal{C}^{B}\times\mathcal{C}^{S}\times\mathcal{C}^{F}\times\mathcal{C}^{M}
$$
It is worth noting that, while our AutoML default search space includes all possible methods, representing the largest possible space, it is flexible and can be modified (add/remove) according to user needs. 

Although the optimization theoretically guarantees finding the global optimum, in practice, it is nearly impossible due to limited computing resources, especially given the multi-dimensional search space we have designed in our AutoML. Thus, we introduce two of the most apparent and natural constraints, \textit{time} and \textit{number of trials} as the computing budget. These two constraints are referred to as \textit{time budget} and \textit{evaluation budget}, respectively, and both can be modified according to user demands. The time budget denotes the maximum allowed runtime for experiments, while the evaluation budget limits the number of evaluations or trials executed during the experiments. Both budgets are audited before each round of evaluations to determine whether a new trial should be generated. A new trial is initiated only if both budgets are not depleted. It is worth noting that the time and evaluation budgets are highly correlated: a larger number of trials typically requires a longer runtime, while allowing for longer experiments generally enables deeper exploration of the search space. Furthermore, since the intermediate optimal loss during the search is non-increasing with respect to the number of trials for fixed sampling procedures, the optimization guarantees at least non-degenerating performance as more time is spent searching or more sets of hyperparameters are explored. Consequently, increasing either the time or evaluation budget generally improves the performance of the optimization in practice.

For each round of evaluation, a specific set of hyperparameters is sampled from the search space $\mathcal{U}$ using a predefined sampling method, commonly referred to as the \textit{Search Algorithm}. The sampled hyperparameter sets can be either independent of each other (e.g., Random Search, Grid Search) or conditional on previous evaluations (e.g., Bayesian Search, \citet{NIPS2012_05311655}, \citet{wu2019hyperparameter}). Each sampled hyperparameter set consists of methods of encoding, imputation, balancing, scaling, feature selection, and an ML model, along with corresponding hyperparameter settings. The initialized pipeline, comprising all method objects, is then fed into the extended objective function $\mathcal{V}$ as inputs to get the evaluation loss at the end of each round. The evaluation losses, which serve as indicators of the pipeline's performance, guide the construction of the optimal pipeline. This optimal pipeline represents the final product of the optimization and modeling task. The optimization is formulated as Algorithm \ref{alg:automl}. Along with the optimal pipeline, all training records, and preprocessed train/test sets are stored for examination if needed.

{\SetAlgoNoLine
\begin{algorithm}
\caption{The AutoML optimization}\label{alg:automl}
\KwIn{Dataset $\mathcal{D}=(\mathcal{D}_{train}, \mathcal{D}_{valid})$; Search space $\mathcal{U}$; Time budget $T$; Evaluation budget $G$; Search algorithm $Samp$}
\KwOut{Optimal pipline with hyperparameter settings $\mathcal{P}^{*}$}
$k = 0$ \Comment*[r]{Round of evaluation}
$t^{re} = T$ \Comment*[r]{Remaining time budget}
$g^{re} = G$ \Comment*[r]{Remaining evaluation budget}
\While{$t^{re} > 0$ and $g^{re} > 0$}{
    $t^{start}=CurrentTime$;\\
    $(E^{(k)}, \lambda_{E}^{(k)}), (I^{(k)}, \lambda_{I}^{(k)}), (B^{(k)}, \lambda_{B}^{(k)}), (S^{(k)}, \lambda_{S}^{(k)}), (F^{(k)}, \lambda_{F}^{(k)}), (M^{(k)}, \lambda_{M}^{(k)})=Samp^{(k)}(\mathcal{U})$;\\
    $\mathcal{P}_{k}=M^{(k)}_{\lambda_{M}^{(k)}}\circ F^{(k)}_{\lambda_{F}^{(k)}}\circ S^{(k)}_{\lambda_{S}^{(k)}}\circ B^{(k)}_{\lambda_{B}^{(k)}}\circ I^{(k)}_{\lambda_{I}^{(k)}}\circ E^{(k)}_{\lambda_{E}^{(k)}}$;\\
    $L^{eval, (k)}=\mathcal{V}(\mathcal{L}, \mathcal{P}_{k}, \mathcal{D})$;\\
    $t^{end}=CurrentTime$;\\
    $k = k + 1$;\\
    $t^{re} = t^{re} - (t^{end} - t^{start})$;\\
    $g^{re} = g^{re} - 1$;\\
    \Comment{Recording hyperparameters, preprocessed data, trained pipeline}
}
$k^{*}=\underset{k}{\mathrm{argmin}}L^{eval, (k)}$ \Comment*[r]{Find optimal pipeline order}
$\mathcal{P}^{*}=\mathcal{P}_{k^{*}}$;\\
\Return{$\mathcal{P}^{*}$};\\ 
\end{algorithm}}

To enable efficient optimization, we employ \textit{Ray Tune} created by \citet{Liaw2018}, a sub-system of the \textit{Ray} library. \textit{Ray} is a flexible and scalable distributed computing framework designed for high-performance and parallel computing tasks, specialized in ML workloads. It simplifies the development of distributed applications, allowing users to parallelize and scale their workloads effortlessly across clusters. Complementing \textit{Ray}, \textit{Ray Tune} is a specialized library for hyperparameter tuning. It provides a comprehensive suite of functionalities for efficient experimentation and optimization of ML models. \textit{Ray Tune} enables users to systematically search through hyperparameter spaces, leveraging state-of-the-art optimization algorithms to fine-tune models for optimal performance. In our AutoML, \textit{Ray Tune} coordinates the optimization experiments by recording key information and facilitating the tuning process. Additionally, we harness two fundamental capabilities of \textit{Ray Tune}: compatibility with mainstream ML model frameworks and seamless integration with a diverse collection of search algorithms. The compatibility of \textit{Ray Tune} with major ML model frameworks like Scikit-Learn \citep{sklearn}, PyTorch \citep{NEURIPS2019_9015}, LightGBM \citep{Ke2017}, XGBoost \citep{Chen2016}, pyGAM \citep{pygam} allows us to utilize models ranging from simple linear models to tree-based ensemble models and deep neural networks. The broad range of available ML models ensures the versatility and applicability of our AutoML to a wide array of ML tasks and datasets. The search algorithms iteratively sample hyperparameter sets from the defined search space. The selection of an appropriate search algorithm significantly influences the attainment of the optimal solution and efficiency. Therefore, search algorithms are crucial to both the efficiency and effectiveness of optimization problems. As summarized by \citet{yang2020hyperparameter}, various search algorithms have been proposed, ranging from fixed algorithms such as grid search to completely random algorithms like random search and algorithms conditioned on previous experiments like evolutionary algorithms \citep{youngOptimizingDeepLearning2015}, gradient-based optimization \citep{bakhteevComprehensiveAnalysisGradientbased2020}. The seamless integration or conversion of frequently-used search algorithms like HyperOpt \citep{Bergstra2013}, Nevergrad \citep{nevergrad}, and Optuna \citep{Akiba2019}, combined with the parallel hyperparameter tuning architecture supported by \textit{Ray Tune}, enables us to create a more flexible environment for experiment settings while ensuring efficiency and effectiveness.

\subsection{Ensemble Model}\label{subsec:ensemble}

As discussed previously, ensemble learning offers alternative solutions to the imbalance problem. In addition to the sampling techniques summarized in Subsection \ref{subsec:pipeline}, we integrate ensemble learning into our AutoML to further address the imbalance problem. Ensemble models combine multiple trained models into a single model by aggregating their predictions, potentially addressing the imbalance problem that individual models cannot effectively handle. Moreover, constructing an ensemble model is recognized as an effective technique for achieving state-of-the-art performance in practical applications. In our AutoML, pipeline-level ensembles are integrated within the pipeline optimization framework. Consequently, the pipelines $\{\mathcal{P}_{k}\}$, $k=1, 2, ..., G$, generated during each evaluation round in Algorithm \ref{alg:automl} can naturally serve as candidates for forming the ensemble model. To enhance the flexibility of the ensemble models, we implement three major ensemble structures: \textit{Stacking}, \textit{Bagging}, and \textit{Boosting}, as summarized in \citet{dong2020survey}.

Considering a total $G$ trained pipelines constrained by the time/evaluation budget following specific training protocols, $H$ pipelines are selected to construct the ensemble model. The final predictions of the ensemble model can be computed as the aggregation of individual predictions through certain \textit{Voting Mechanisms}. Consequently, the ensemble model $\Sigma$, given $G$ pipelines $\{\mathcal{P}_{g}\}$, $g=1, 2, \ldots, G$, can be expressed as
\begin{align*}
    \Sigma_{H}&=\Sigma_{H}(\mathcal{P}_{1}, \mathcal{P}_{2}, \ldots, \mathcal{P}_{G})\\
    &=\Sigma_{H}(\mathcal{P}_{(1)}, \mathcal{P}_{(2)}, \ldots, \mathcal{P}_{(H)})
\end{align*}
where $H$ denotes the pre-fixed hyperparameter for the size of the ensemble model (number of selected pipelines), and $\mathcal{P}_{(h)}$ refers to the $h$-th pipeline ranked by the evaluation loss $L^{eval}$ computed from the objective function $\mathcal{V}$. The predictions, given the ensemble model $\Sigma_{H}$ and a input matrix $\textbf{X}$, can be expressed as
\begin{align*}
    \hat{\textbf{y}}&=\Sigma_{H}(\textbf{X})\\
    &=\gamma(\mathcal{P}_{(1)}(\textbf{X}), \mathcal{P}_{(2)}(\textbf{X}), \ldots, \mathcal{P}_{(H)}(\textbf{X}))
\end{align*}
where $\gamma$ denotes the voting mechanism attached to the ensemble model $\Sigma_{H}$.

It is important to note that the three ensemble structures only function as the protocol of the training diagram, predictions through the voting mechanism, and, in our work, validation for the completion of the task. They do not involve any additional training procedures. The optimization of each individual pipeline given the input training/validation datasets thus remains unaffected by the deployment of ensemble models. Specifically, the naming of these ensemble models distinguishes them by their training diagrams. 
The stacking ensemble models utilize a fully parallel optimization approach across entire datasets, where multiple base pipelines are trained independently and their predictions are aggregated through the voting mechanism. In contrast, bagging ensemble models are optimized by training multiple base pipelines on random subsets of the data, aiming to reduce variance and enhance generalization. Boosting ensemble models iteratively improve model predictions by focusing on the residuals from previous pipelines, sequentially refining predictions to minimize overall error. Reference Appendix \ref{appendix_sec:enc} for details of the ensemble strategies.

\subsection{Loss Functions}\label{subsec:lossfunction}

Loss functions play a pivotal role in both model creation and evaluation. By mapping pairs of true observations and model predictions to a real value, loss functions provide a quantifiable measure of model performance. As summarized by \citet{wangComprehensiveSurveyLoss2022}, a variety of loss functions have been developed to address the specific demands of different ML tasks. These loss functions are meticulously designed to optimize the model performance across diverse applications, ensuring alignment with the objectives and constraints of each particular task.

The choice of loss functions can significantly influence model creation and, ultimately, the success of ML models. In financial modeling, \citet{bamsLossFunctionsOption2009} demonstrate that the choice of loss functions has a substantial impact on option valuation, underscoring their critical role in this domain. Similarly, in deep learning, the design of appropriate loss functions is crucial for tasks such as Object Detection \citep{linFocalLossDense2017} and Image Segmentation \citep{salehiTverskyLossFunction2017}. In the insurance domain, one of the major challenges is the imbalanced distribution of response variables, which complicates the direct application of ML models. To address this issue, researchers have proposed carefully calibrated imbalance learning algorithms and adjusted cost-sensitive loss functions, as highlighted by \citet{huImbalancedLearningInsurance2022a}, \citet{zhangBayesianCARTModels2024}, and \citet{soSAMMEC2Algorithm2024}.

Choosing the appropriate loss function is essential for aligning ML models with insurance business objectives. It ensures that models not only perform well technically, but also deliver outcomes that are meaningful and beneficial to the insurance business. Loss functions guide the model during training by quantifying errors, ensuring that the model learns to minimize the types of errors that matter the most to the insurance context. Our AutoML framework provides a variety of common loss functions suitable for both imbalanced and balanced learning scenarios. Additionally, it offers flexibility for users to employ customized loss functions based on their specific needs. This customization allows users to define loss functions that better capture their unique objectives, leading to more relevant and actionable insights. For instance, in sensitive areas like insurance pricing, ensuring fairness across different demographic groups is crucial. Custom loss functions can be designed to enforce fairness constraints and comply with regulatory requirements, helping to ensure ethical and equitable outcomes.


\section{AutoML in Action}\label{sec:exp}

To demonstrate the feasibility and efficacy of our AutoML, we conducted experiments using several datasets studied by actuarial science researchers. We evaluated both the performance and run-time of these experiments. Our AutoML framework is user-friendly and requires only a few lines of code to deploy, making it accessible to inexperienced users. Please refer to Appendix \ref{appendix_sec:code} for the code required to run the experiments, along with the corresponding descriptions.

\subsection{French Motor Third-Part Liability}\label{subsec:freMTPL2freq}

In this experiment, we use the French Motor Third-Part Liability datasets, \textit{freMTPL2freq}, from package \textit{CASDatasets} \citep{charpentier2014computational}, which comprises 677,991 motor liability policies collected in France with the response variable \textit{ClaimNb}, indicating number of claims during the exposure period and 10 categorical/numerical explanatory variables (excluding the policy ID, \textit{IDpol}). Refer to Table \ref{tab:freMTPL_data} for the full list of features and response variables and their descriptions. The task is to predict future claim frequency, which is framed as a regression problem. We follow the same random seed and train/test split percentage suggested by \citet{noll2020case} to replicate the train/test sets, without applying any preprocessing techniques before feeding the data into the AutoML pipeline. The performance of the experiment is evaluated by mean Poisson deviance, as the same metrics utilized in \citet{noll2020case}, which can be expressed as 
$$
\mathcal{L}_{Poi}(\textbf{y}, \hat{\textbf{y}})=\dfrac{2}{Z}\sum_{z=1}^{Z}(\hat{y}_{z}-y_{z}+y_{z}\log(\dfrac{y_{z}}{\hat{y}_{z}}))
$$
for a total of $Z$ true response values $y_{z}$ and predictions $\hat{y}_{z}$ ($z=1,2,...,Z$). The evaluation of mean Poisson deviance is common in actuarial practice for claim frequency modeling. 

\begin{table}[!ht]
\centering
\begin{tabular}{c c c l} \hline
Category & Name & Type & Description  \\
\hline
\multirow{10}*{Features} & \textit{Area} & Categorical & Density values of the community \\[-0.1ex]
& \textit{BonusMalus} & Numerical & Bonus/Malus of the driver \\[-0.1ex]
& \textit{Density} & Numerical & Density of inhabitants in the neighborhood \\[-0.1ex]
& \textit{DrivAge} & Numerical & Age of the driver \\[-0.1ex]
& \textit{Exposure} & Numerical & Exposure length of enforced policies \\[-0.1ex]
& \textit{Region} & Categorical & Region of the policies by categories \\[-0.1ex]
& \textit{VehAge} & Numerical & Age of the vehicle \\[-0.1ex]
& \textit{VehBrand} & Categorical & Brand of the vehicle \\[-0.1ex]
& \textit{VehGas} & Categorical & Gasoline type of the vehicle \\[-0.1ex]
& \textit{VehPower} & Numerical & Power of the vehicle \\[-0.1ex]
\cmidrule{2-4}
Response & \textit{ClaimNb} & Numerical & Number of claims reported \\[-0.1ex]
\hline
\end{tabular}
\caption{Features \& Response variables of freMTPL2freq dataset}
\label{tab:freMTPL_data}
\end{table}

\begin{table}[!ht]
\centering
\begin{tabular}{c c c c c c} \hline
G & T/s & runtime/s & Train Deviance & Test Deviance \\ 
\hline
8 & 900 & 807.62 & 0.3622 & 0.3689 \\[-0.1ex]
16 & 1,800 & 1,082.21 & 0.3826 & 0.3890 \\[-0.1ex]
32 & 3,600 & 2,092.15 & 0.3156 &  0.3250 \\[-0.1ex]
64 & 7,200 & 4,417.51 & 0.3022 &  0.3122  \\[-0.1ex]
128 & 14,400 & 8,052.91 & 0.2925 & 0.3034 \\[-0.1ex]
256 & 28,800 & 12,624.60 & 0.2779 & \textbf{0.3009} \\[-0.1ex]
512 & 57,600 & 34,036.03 & 0.2762 & 0.3020 \\[-0.1ex]
1024 & 115,200 & 63,401.81 & \textbf{0.2539} & 0.3114 \\[-0.1ex]
\hline
\end{tabular}
\caption{AutoML performance on freMTPL2freq dataset}
\label{tab:freMTPL_result}
\end{table}

\begin{figure}[htbp]
\centering
\includegraphics[width=0.8\linewidth]{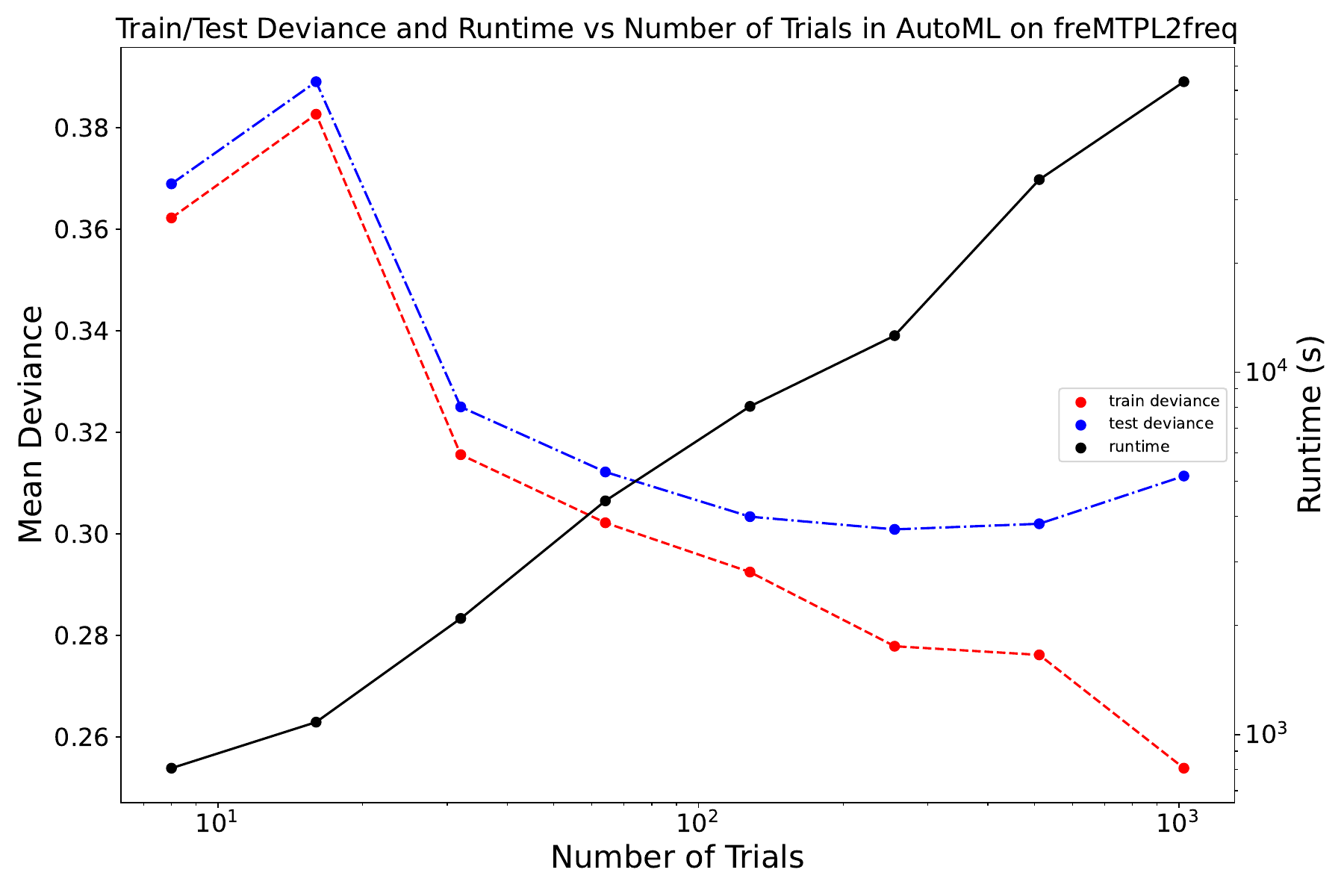}
\caption{Train/Test deviance and runtime on freMTPL2freq dataset}
\label{fig:freMTPL2freq}
\end{figure}

To illustrate the performance of our AutoML in terms of evaluation and time budget, we train the AutoML across a range of increasing evaluation budget $G$ and time budget $T$. The runtime and train/test deviance for each ensemble of fitted pipelines are shown in Table \ref{tab:freMTPL_result} with a visualization provided in Figure \ref{fig:freMTPL2freq} on a log scale. In Figure \ref{fig:freMTPL2freq}, it is evident that as $G$ and $T$ increase synchronously, the runtime increases approximately linearly with the evaluation budget. Concurrently, train/test deviance decreases, indicating the performance gain from a more extensive hyperparameter search. The superior performance at $G=8$ compared to $G=16$ can be explained by the insufficient number of fitted pipelines at early training stages. At the starting point $G=8$, only two valid pipelines could be selected to form the ensemble model out of five trained pipelines, due to the metric of mean Poisson deviance only allowing non-zero predictions. In contrast, at $G=16$, although five valid pipelines could be selected under the current evaluation and time budget, the ensemble construction is underperforming compared to the one from the $G=8$ stage. This is because the early stages of hyperparameter search do not provide enough valid and high-performing pipelines. This suggests that users should consider increasing the evaluation and time budget for better results. On the other hand, the increasing test deviance observed from $G=256$ to $G=1,024$ does not necessarily indicate over-fitting but rather reflects the limitations of the search concerning the vast hyperparameter space designed for the AutoML pipeline. This suggests that users should consider stopping the AutoML process and investigating the fitted pipelines. 

\subsection{Wisconsin Local Government Property Insurance Fund}\label{subsec:LGPIF}

The Wisconsin Local Government Property Insurance Fund dataset, \textit{LGPIF}, introduced in \citet{frees2016multivariate}, is another classical real-world dataset for actuarial science researchers. The local government property insurance covers six different groups: building and content (BC), contractor’s equipment (IM), comprehensive new (PN), comprehensive old (PO), collision new (CN), and collision old (CO). The dataset consists of 5,677 policies collected from the years 2006 to 2010 as a train set and 1,098 unique policies of the year 2011 as a test set. In our AutoML experiments, following \citet{quan2018predictive}, we select only one line of business, and variable \textit{yAvgBC}, the average claim sizes of coverage Building and Content, as the response, making it a supervised regression task. The dataset includes 8 continuous variables describing coverage and deductible information and 13 categorical indicators of insured entities and history claims are utilized as features. For details on the variables and their descriptions, refer to Table \ref{tab:LGPIF_data}. Specifically, as instructed in \citet{quan2018predictive}, we apply the logarithmic transformation on response variable, \textit{yAvgBC}, and 6 coverage variables, which can be expressed as
$$
\hat{x}=\log (x + 1)
$$
where $x$ denotes the original variable and $\hat{x}$ refers to the transformed variable. In training our AutoML, we optimize based on the loss function using the Coefficient of Determination ($R^{2}$). For a total of $Z$ true values $\textbf{y}$ and their corresponding predictions $\hat{\textbf{y}}$ generated by fitted AutoML, can be written as
$$
\mathcal{L}_{R^{2}}(\textbf{y}, \hat{\textbf{y}})=\dfrac{RSS}{TSS}
$$
where $RSS=\sum_{z=1}^{Z}(\hat{y}_{z}-y_{z})^{2}$ and $TSS=\sum_{z=1}^{Z}(y_{z}-\bar{\textbf{y}})^{2}$ denote the regression sum of squares and total sum of squares respectively and $\bar{\textbf{y}}=\dfrac{1}{Z}\sum_{z=1}^{Z}y_{z}$ is the mean of all true values.

\begin{table}[!ht]
\centering
\begin{tabular}{c c c p{5.5cm}} 
\hline
Category & Name & Type & Description  \\
\hline
\multirow{12}*{Feature} & \textit{Type} & Categorical & Binary indicator of property type (City, County, Misc, School, Town, Village) \\[-0.1ex]
& \textit{IsRC} & Categorical & Binary indicator of replacement cost \\[-0.1ex]
& \textit{Coverage} & Numerical & Coverage of the property (BC, IM, PN, PO, CN, CO) \\[-0.1ex]
& \textit{lnDeduct} & Numerical & Logarithm of deductible (BC, IM) \\[-0.1ex]
& \textit{NoClaimCredit} & Categorical & Binary indicator of prior claim reports (BC, IM, PN, PO, CN, CO) \\[-0.1ex]
\cmidrule{2-4}
Response & \textit{yAvgBC} & Numerical & Average claim sizes \\[-0.1ex]
\hline
\end{tabular}
\caption{Features \& Response variables of LGPIF dataset}
\label{tab:LGPIF_data}
\end{table}

\begin{table}[!ht]
\centering
\begin{tabular}{c c c c c c} \hline
G & T/s & runtime/s & Train $R^{2}$ & Test $R^{2}$ \\ 
\hline
8 & 900 & 1202.55 & 0.2267 & 0.2151 \\[-0.1ex]
16 & 1,800 & 1,533.87 & 0.2700 & 0.2268 \\[-0.1ex]
32 & 3,600 & 1,513.57 & 0.2373 & 0.2235 \\[-0.1ex]
64 & 7,200 & 2,891.36 & 0.2674 & 0.2255 \\[-0.1ex]
128 & 14,400 & 3,367.43 & 0.3409 & 0.2361 \\[-0.1ex]
256 & 28,800 & 8,413.55 & 0.3330 & 0.2360 \\[-0.1ex]
512 & 57,600 & 10,313.03 & \textbf{0.3424} & 0.2372 \\[-0.1ex]
1024 & 115,200 & 15,282.33 & 0.3197 & \textbf{0.2377} \\[-0.1ex]
2048 & 230,400 & 40,856.35 & 0.3260 & 0.2360 \\[-0.1ex]
\hline
\end{tabular}
\caption{AutoML performance on LGPIF dataset}
\label{tab:LGPIF_result}
\end{table}

\begin{figure}[htbp]
\centering
\includegraphics[width=0.8\linewidth]{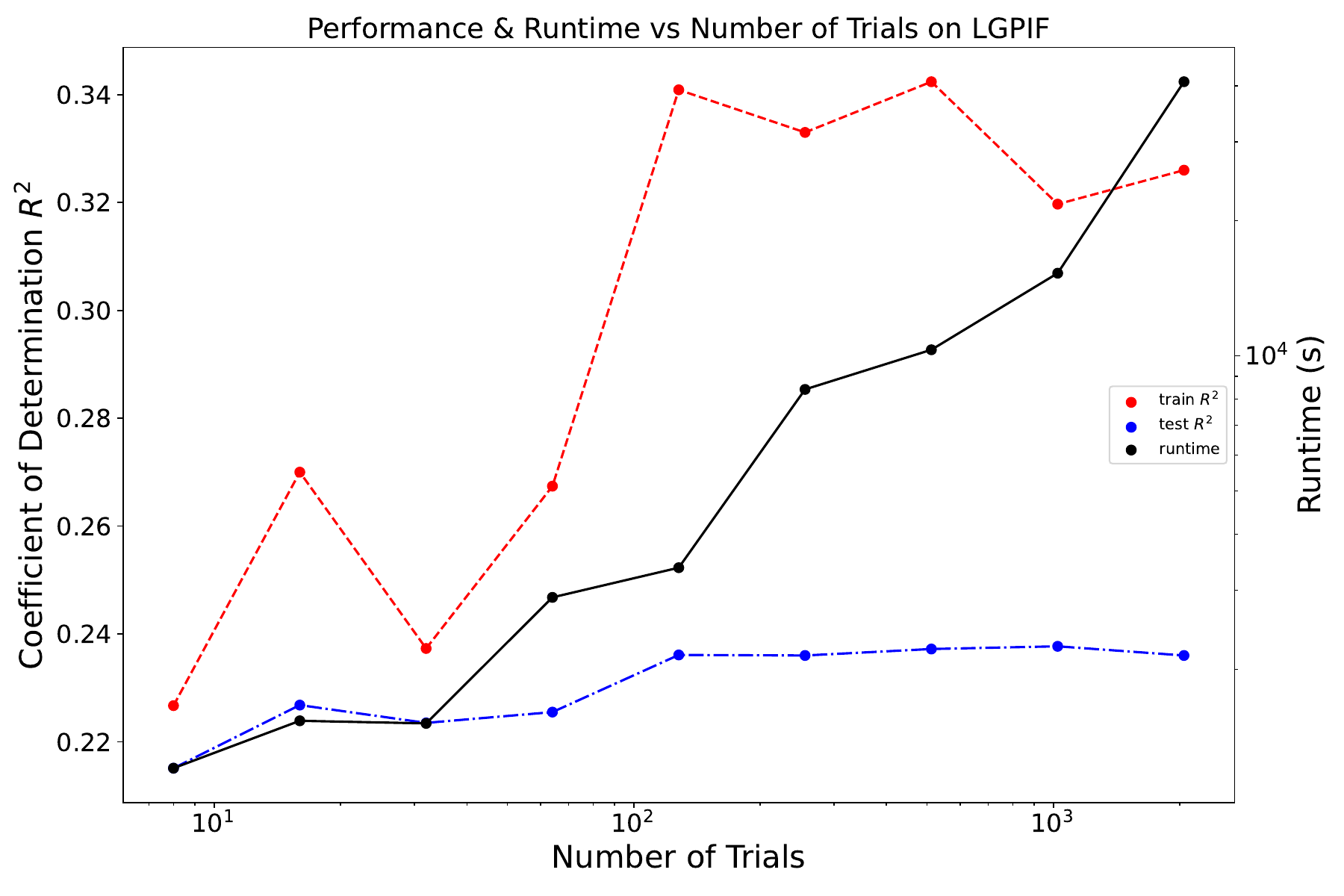}
\caption{Train/Test error and runtime on LGPIF dataset}
\label{fig:LGPIF}
\end{figure}

The experiment follows the same structure as demonstrated in Subsection \ref{subsec:freMTPL2freq}, where we record the runtime and train/test errors as the evaluation and time budget increase. Table \ref{tab:LGPIF_result} and Figure \ref{fig:LGPIF} summarize the results and the visualization correspondingly. Both Table \ref{tab:LGPIF_result} and Figure \ref{fig:LGPIF} suggest that scaling evaluation and time budget on the \textit{LGPIF} dataset has a less significant impact on the performance compared to \textit{freMTPL2freq} dataset. One obvious reason is that the \textit{LGPIF} dataset is significantly smaller than the \textit{freMTPL2freq} dataset, requiring fewer trials to achieve suitable results from our AutoML. In addition, we also notice that, throughout the experiments, the best-performing individual pipelines constantly demonstrate improvement on the train set. This outcome validates the feasibility of substituting manual tuning with automatic optimization, showcasing the efficacy of our approach. 

\subsection{Australian Automobile Insurance}\label{subsec:ausprivauto}

The Automobile claim datasets in Australia, \textit{ausprivauto}, has multiple response variables related to insurance claims introduced by \citet{de2008generalized} and also collected in the package \textit{CASDatasets} \citep{charpentier2014computational}. For our experiments, we use subset \textit{autoprivauto0405}, which consists of 67,856 one-year auto insurance policies enforced in the year 2004 or 2005. We select the variable \textit{ClaimOcc}, a binary indicator of the occurrence of claim events, to test out the performance of AutoML on classification tasks. To differentiate this experiment from other experiments on the \textit{ausprivauto} dataset, we denote the binary claim occurrence estimation task as \textit{ausprivauto\_occ}. The features comprise 4 categorical variables, describing the vehicle age groups (\textit{VechAge}), vehicle body groups (\textit{VehBody}), drivers' gender (\textit{Gender}), and drivers' age groups (\textit{DrivAge}), and two continuous features of vehicles' value (\textit{VehValue}) and exposure (\textit{Exposure}). Following the example of applying Logistic Regression in \citet{de2008generalized}, our AutoML can achieve the training AUC score of 0.9401 by training on the entire dataset, with 128 rounds of evaluation and 2,047.21 seconds of fitting. Compared to the training AUC score of 0.662 achieved by the exposure-adjusted model \citep{de2008generalized}, our AutoML fitted training dataset is better than the Generalized Linear Model (GLM) family. However, the high AUC score may also indicate over-fitting, given that no cross-validation is applied. To ensure robust performance, subsequent experiments are conducted with the same feature sets and same loss function but with a 90/10 train/test split ratio and 4-fold cross-validation, following the experimental design suggested by \citet{siAutomobileInsuranceClaim2022}. The loss function, AUC, can be written as
$$
\mathcal{L}_{AUC}(\textbf{y}, \hat{\textbf{y}}_{prob})=\dfrac{\sum_{\mu: y_{\mu}=0}\sum_{\nu: y_{\nu}=1}\mathbbm{1}_{\hat{y}_{prob, \mu}<\hat{y}_{prob, \nu}}}{(\sum_{\mu=1}^{Z}\mathbbm{1}_{y_{\mu}=0})(\sum_{\nu=1}^{Z}\mathbbm{1}_{y_{\nu}=1})}
$$
for a binary classification problem, where $\textbf{y}$ and $\hat{\textbf{y}}_{prob}$ are the true classes and predicted probabilities of the positive class.

\begin{table}[!ht]
\centering
\begin{tabular}{c c c c c c} \hline
G & T/s & runtime/s & Train AUC & Test AUC \\ 
\hline
8 & 900 & 565.15 & 0.8681 & 0.5407 \\[-0.1ex]
16 & 1,800 & 629.83 & 0.8489 & 0.6253 \\[-0.1ex]
32 & 3,600 & 1,127.55 & 0.6578 & 0.6754 \\[-0.1ex]
64 & 7,200 & 2,506.17 & 0.6560 & 0.6739 \\[-0.1ex]
128 & 14,400 & 3,749.68 & 0.6576 & 0.6754 \\[-0.1ex]
256 & 28,800 & 4,598.22 & 0.6600 & 0.6770 \\[-0.1ex]
512 & 57,600 & 9,709.30 & 0.6602 & 0.6774 \\[-0.1ex]
1024 & 115,200 & 18,938.53 & 0.6609 & 0.6815 \\[-0.1ex]
2048 & 230,400 & 30,951.82 & 0.6626 & \textbf{0.6831} \\[-0.1ex]
4096 & 460,800 & 79,438.58 & \textbf{0.6627} & 0.6831 \\[-0.1ex]
\hline
\end{tabular}
\caption{AutoML performance on ausprivauto claim occurrence}
\label{tab:ausprivauto_occ_result}
\end{table}

\begin{figure}[htbp]
\centering
\includegraphics[width=0.8\linewidth]{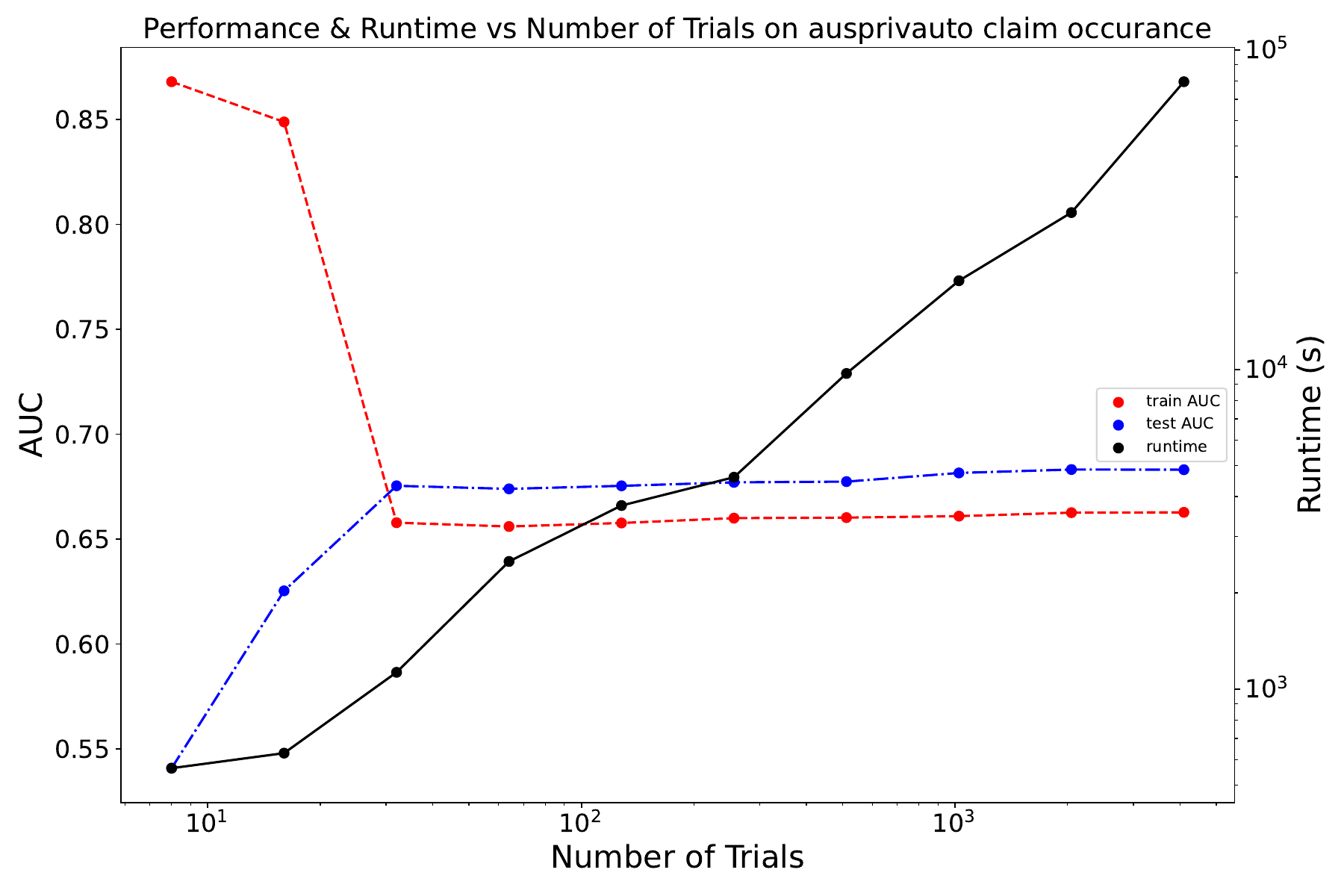}
\caption{Train/Test error and runtime on ausprivauto occurrence dataset}
\label{fig:ausprivauto_occ}
\end{figure}

The numerical runtime and performance and its corresponding visualization can be found in Table \ref{tab:ausprivauto_occ_result} and Figure \ref{fig:ausprivauto_occ} respectively. As observed, the runtime exhibits a similar scaling pattern as seen in the behaviors of the \textit{freMTPL2freq} and \textit{LGPIF} datasets. Experiments with an evaluation budget $G \leq 16$ show a notable discrepancy between the train and test metrics compared to the other entries in the table. For experiments with $G \leq 16$, the best-performing model architecture is the kNN. Obviously, the kNN model tends to be overfitting in this dataset and leads to a disparity between training and testing metrics. This disparity is significantly mitigated for experiments with $G \geq 32$, as the instability of the kNN architecture during k-fold cross-validation leads to the selection of other model architectures. Both train and test performance metrics steadily improve with the increasing evaluation budget and time budget.

\begin{figure}[htbp]
\centering
\includegraphics[width=0.8\linewidth]{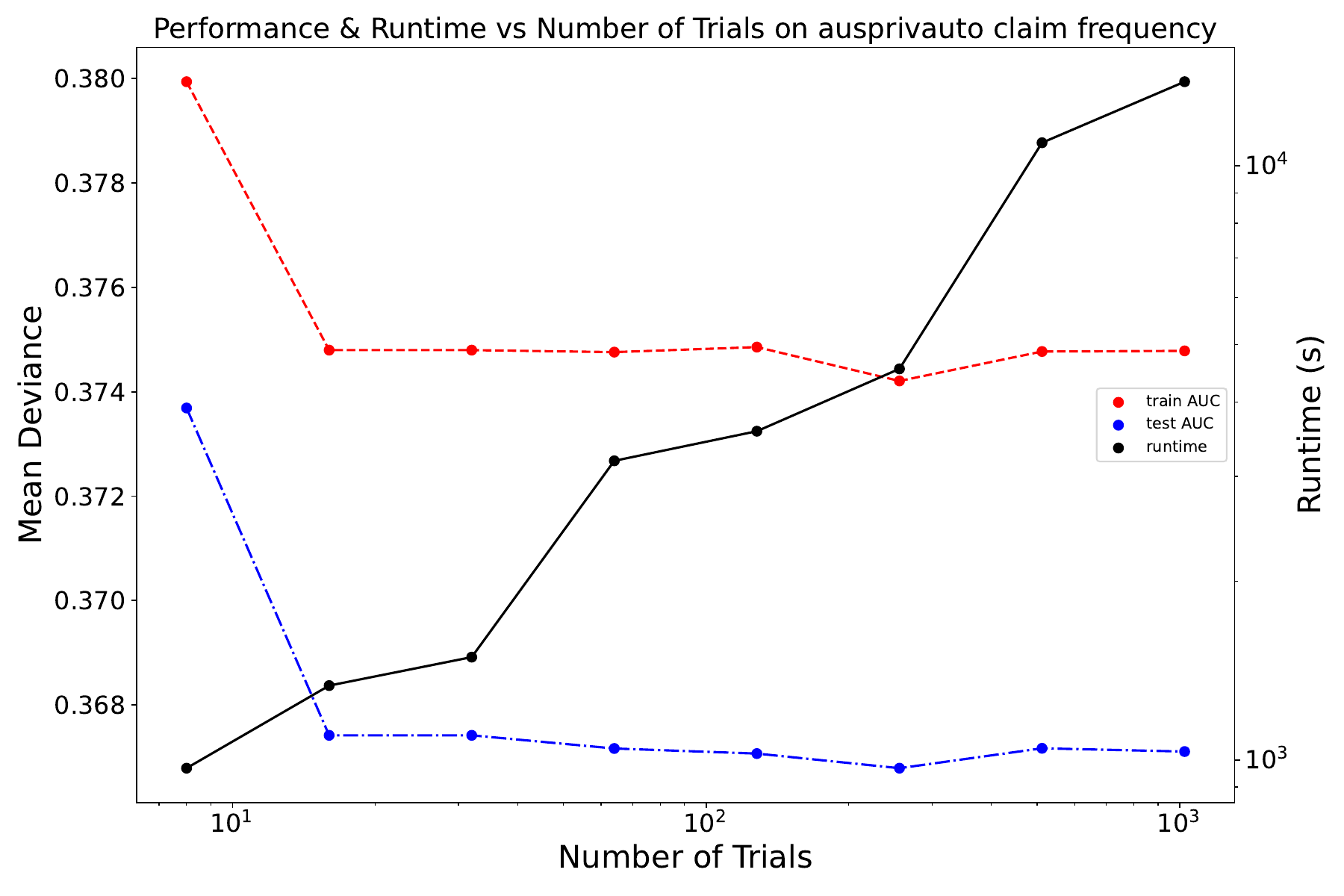}
\caption{Train/Test deviance and runtime on ausprivauto frequency dataset}
\label{fig:ausprivauto_fre}
\end{figure}

\begin{figure}[htbp]
\centering
\includegraphics[width=0.8\linewidth]{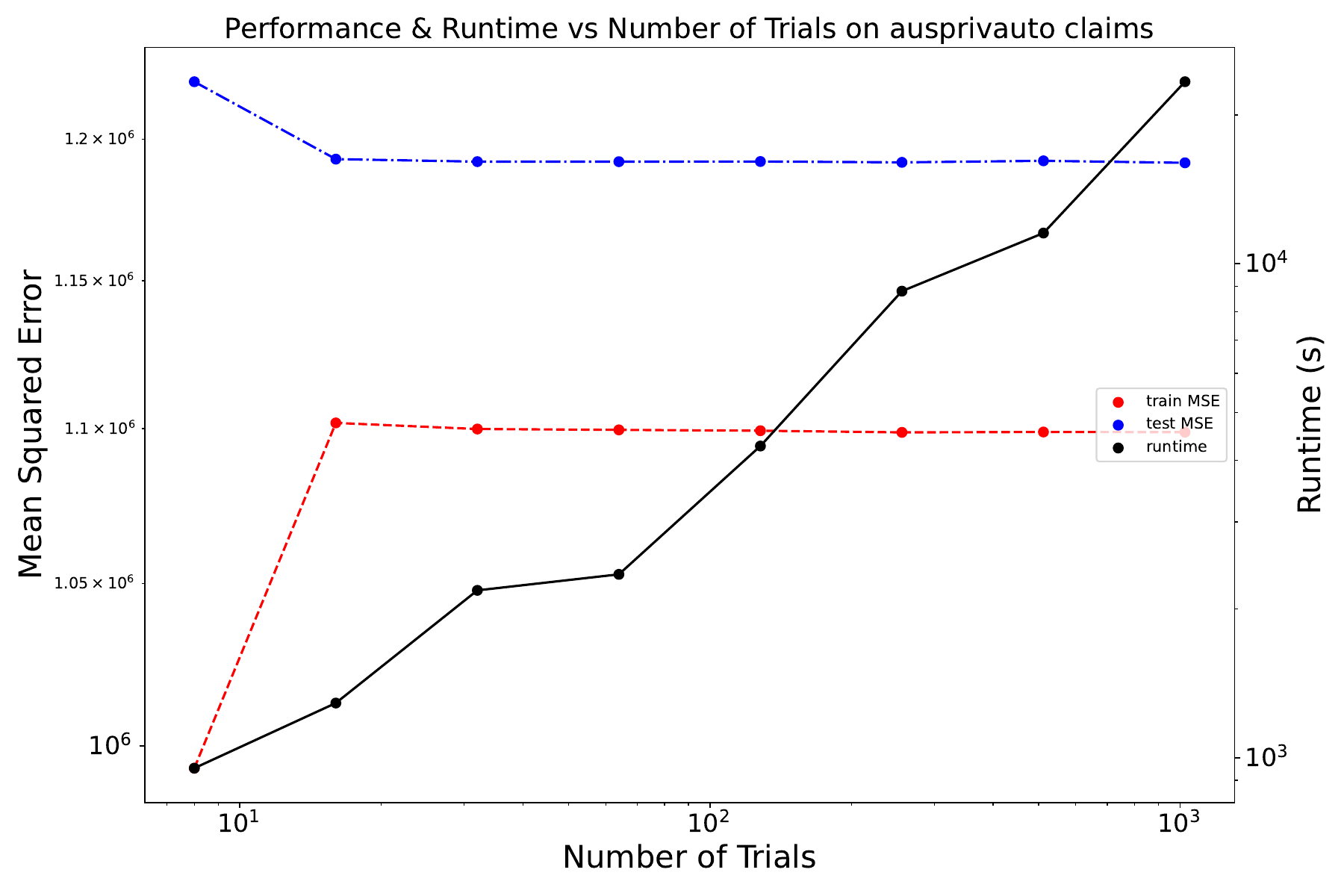}
\caption{Train/Test deviance and runtime on ausprivauto claims}
\label{fig:ausprivauto}
\end{figure}

Additionally, we can construct two regression models using the response variables for the number of claims reported, \textit{ClaimNb}, and for the aggregated claim amounts, \textit{ClaimAmount}, with the same feature sets. In the following discussion, we refer to the experiments conducted on claim frequency and claim amounts as \textit{ausprivauto\_fre} and \textit{ausprivauto}, respectively. The AutoML optimization in these two regression experiments tries to minimize the mean Poisson deviance $\mathcal{L}_{Poi}$ and Mean Squared Error (MSE) between the true values $\textbf{y}$ and predicted values $\hat{\textbf{y}}$ respectively. The metric of MSE is defined as:
$$
\mathcal{L}_{MSE}(\textbf{y}, \hat{\textbf{y}})=\dfrac{1}{Z}\sum_{z=1}^{Z}(y_{z}-\hat{y}_{z})^{2}
$$
The evaluation performance and runtime for \textit{ausprivauto\_fre} and \textit{ausprivauto} datasets are summarized in Table \ref{tab:ausprivauto_fre_result} and Table \ref{tab:ausprivauto_result}, and the corresponding visualization plots are illustrated as Figure \ref{fig:ausprivauto_fre} and Figure \ref{fig:ausprivauto}. As the tables and plots illustrate, our AutoML exhibits a diminishing trend in the improvement of performance metrics over time and evaluation budget. 

\begin{table}[!ht]
\centering
\begin{tabular}{c c c c c c} \hline
G & T/s & runtime/s & Train Deviance & Test Deviance \\ 
\hline
8 & 900 & 968.38 & 0.3799 & 0.3737 \\[-0.1ex]
16 & 1,800 & 1,333.62 & 0.3748 & 0.3674 \\[-0.1ex]
32 & 3,600 & 1,489.29 & 0.3748 & 0.3674 \\[-0.1ex]
64 & 7,200 & 3,187.05 & 0.3748 & 0.3672  \\[-0.1ex]
128 & 14,400 & 3,574,53 & 0.3749 & 0.3670 \\[-0.1ex]
256 & 28,800 & 4,551.89 & \textbf{0.3742} & \textbf{0.3668} \\[-0.1ex]
512 & 57,600 & 10,933.49 & 0.3748 & 0.3672 \\[-0.1ex]
1024 & 115,200 & 13,849.67 & 0.3748 & 0.3671 \\[-0.1ex]
\hline
\end{tabular}
\caption{AutoML performance on ausprivauto claim frequency}
\label{tab:ausprivauto_fre_result}
\end{table}

\begin{table}[!ht]
\centering
\begin{tabular}{c c c c c c} \hline
G & T/s & runtime/s & Train MSE & Test MSE \\ 
\hline
8 & 900 & 952.30 & \textbf{993,285.74} & 1,220,835.99 \\[-0.1ex]
16 & 1,800 & 1,290.22 & 1,101,871.13 & 1,192,769.13 \\[-0.1ex]
32 & 3,600 & 2,181.53 & 1,099,855.88 & 1,191,872.59 \\[-0.1ex]
64 & 7,200 & 2,350.45 & 1,099,573.20 & 1,191,859.57 \\[-0.1ex]
128 & 14,400 & 4,274,43 & 1,099,296.76 & 1,191,913.62 \\[-0.1ex]
256 & 28,800 & 8,797.69 & 1,098,746.18 & 1,191,618.57 \\[-0.1ex]
512 & 57,600 & 11,533.74 & 1,098,873.21 & 1,192,162.05 \\[-0.1ex]
1024 & 115,200 & 23,347.84 & 1,098,817.89 & \textbf{1,191,450,96} \\[-0.1ex]
\hline
\end{tabular}
\caption{AutoML performance on ausprivauto claim amount}
\label{tab:ausprivauto_result}
\end{table}

\subsection{AutoML as a Benchmark}

To demonstrate the potential of our AutoML pipeline as a benchmark, we compare the model performance derived from our AutoML with that of the widely used GLM in insurance pricing, and other actuarial literature studying the same dataset.

To ensure a fair comparison, we build GLM using the same datasets mentioned in the previous experiments. For claim frequency estimation on the \textit{freMTPL2freq} dataset and the \textit{ausprivauto} dataset (i.e., experiment \textit{ausprivauto\_fre}), we employ Poisson regression with the logarithm of exposure as the offset. For claim occurrence on \textit{ausprivauto} (i.e., experiment \textit{ausprivauto\_occ}), we utilize the Logistic regression with a Bernoulli distribution. Lastly, for estimation of aggregated claim amounts on \textit{LGPIF} dataset and \textit{ausprivauto} dataset (i.e., experiment \textit{ausprivauto}), our model is based on GLM with a Tweedie family. 

As shown in Table \ref{tab:optimal_GLM}, our AutoML consistently demonstrates superior performance compared to the GLM family under reasonably constrained computational hardware and time budgets. This comparison highlights the effectiveness of the AutoML approach in delivering superior results across various insurance pricing models, highlighting its adaptability and advantages over traditional GLM frameworks.

\begin{table}[!ht]
\small
\centering
\begin{tabular}{c c c c c c } \hline
Data & \multicolumn{3}{c}{GLM Results} & \multicolumn{2}{c}{AutoML}  \\
& Family & Metric & Test loss & G & Test loss \\ 
\hline
\textit{freMTPL2freq} & Poisson & Poisson Deviance & 0.3595 & 256 & \textbf{0.3009} \\[-0.1ex]
\hline
\multirow{5}*{\textit{LGPIF}} & \multirow{5}*{Tweedie} & $R^{2}$ & 0.2062 & \multirow{5}*{1024} & \textbf{0.2377} \\[-0.1ex]
& & Gini & 0.4089 &  & \textbf{0.4187} \\[-0.1ex]
& & ME & 0.1609 &  & \textbf{0.0476} \\[-0.1ex]
& & MSE & 14.0533 &  & \textbf{13.4956} \\[-0.1ex]
& & MAE & \textbf{2.8749} &  & 2.8955 \\[-0.1ex]
\hline
\textit{ausprivauto\_occ} & Bernoulli & AUC & 0.6792 & 2048& \textbf{0.6831} \\[-0.1ex]
\hline
\textit{ausprivauto\_fre} & Poisson & Poisson Deviance & 0.4437 & 256 & \textbf{0.3668} \\[-0.1ex]
\hline
\textit{ausprivauto} & Tweedie & RMSE & 1,091.6741 & 1024 & \textbf{1,091.5361} \\[-0.1ex]
\hline
\end{tabular}
\caption{Comparison of AutoML with GLM models}
\label{tab:optimal_GLM}
\end{table}

Furthermore, to assess how our AutoML performs relative to cutting-edge actuarial research, we gather the performance metrics from AutoML experiments conducted on \textit{freMTPL2freq}, \textit{LGPIF} and claim occurrence in \textit{ausprivauto\_occ} datasets, applying the consistent data-splitting strategies as those discussed in actuarial studies. 

As summarized in Table \ref{tab:optimal} and \ref{tab:freMTPL_result}, AutoML achieves a test Poisson deviance less than 0.3122, with the best performance reaching 0.3009, when the evaluation budget $G$ greater than 64. This result surpasses the test Poisson deviance 0.3149 reported by \citet{wuthrich2019generalized}. Notably, the results provided by researchers are superior to those from a naive Poisson GLM, which has a test Poisson deviance of 0.3595, reflecting the benefits of careful data preprocessing and model selection. However, AutoML consistently outperforms the claim frequency modeling results provided by humans for the \textit{freMTPL2freq} dataset. This suggests the potential for AutoML to eliminate the need for manual data preprocessing and model selection. At the very least, the training results from the AutoML pipelines can inspire humans to discover better solutions.

Additionally, as evident from the \textit{LGPIF} section of Table \ref{tab:optimal}, the optimal predictions generated by our AutoML outperform all metrics except MAE when compared to the results from \citet{quan2018predictive} on the claim severity prediction. While \citet{quan2018predictive} utilized various tree-based models, our AutoML ensemble model leverages multiple state-of-the-art algorithms, resulting in superior prediction performance. A similar trend is observed for the claim occurrence prediction on the \textit{ausprivauto\_occ} dataset, where our AutoML outperforms the stacking ensemble model approach suggested by  \citet{siAutomobileInsuranceClaim2022}. 
The comparison of the experiment \textit{LGPIF} in Table \ref{tab:optimal_GLM} and Table \ref{tab:optimal} suggests that our AutoML not only outperforms the GLM but also has the potential to surpass state-of-the-art ML models developed by human experts.

\begin{table}[!ht]
\small
\centering
\begin{tabular}{c c c c c c } \hline
Data & \multicolumn{3}{c}{Actuarial literature} & \multicolumn{2}{c}{AutoML}  \\
& Source & Metric & Test loss & G & Test loss \\ 
\hline
\textit{freMTPL2freq} & \citet{wuthrich2019generalized} & Poisson Deviance & 0.3149 & 256 & \textbf{0.3009} \\[-0.1ex]
\hline
\multirow{5}*{\textit{LGPIF}} & \multirow{5}*{\citet{quan2018predictive}} & $R^{2}$ & 0.229 & \multirow{5}*{1024} & \textbf{0.2377} \\[-0.1ex]
& & Gini & 0.414 &  & \textbf{0.4187} \\[-0.1ex]
& & ME & 0.048 &  & \textbf{0.0476} \\[-0.1ex]
& & MSE & 13.651 &  & \textbf{13.4956} \\[-0.1ex]
& & MAE & \textbf{2.883} &  & 2.8955 \\[-0.1ex]
\hline
\textit{ausprivauto\_occ} & \citet{siAutomobileInsuranceClaim2022} & AUC & 0.660 & 2048 & \textbf{0.6831} \\[-0.1ex]
\hline
\end{tabular}
\caption{Comparison of AutoML with other actuarial literature}
\label{tab:optimal}
\end{table}

As demonstrated in the comparative analysis presented in Table \ref{tab:optimal_GLM} with the GLM model framework and Table \ref{tab:optimal} alongside selected actuarial literature, our AutoML consistently outperforms in various insurance tasks, spanning both classification and regression. In the context of the insurance sector, our approach encompasses tasks related to predicting claim occurrence, claim frequency, and claim amounts. This performance's superiority across diverse tasks underscores the practicality and effectiveness of our automated approach in the insurance domain.

\section{Conclusion}\label{sec:conclusion}

In conclusion, there is a noticeable gap between insurance practitioners and the data science community when it comes to effectively optimizing and applying ML solutions for both industrial and research purposes. 

Our AutoML effectively bridges this gap by automating the ML pipeline for insurance-related tasks. This framework not only streamlines the modeling process but also demonstrates adaptability and robust performance across various insurance applications. Our comprehensive experimentation and comparison with traditional modeling frameworks, such as GLM, and existing actuarial literature confirm the feasibility and effectiveness of our AutoML approach.

The runtime scalability of our AutoML, as observed in the previous experiments, generally meets expectations, showing a roughly linear increase with the evaluation budget. However, performance improvements may plateau over time or fail to scale significantly, potentially leading to wasted computation and increased energy consumption. Therefore, to attain optimal performance, strategies should include not only increasing the computational budget—through enhanced hardware or extended evaluation/time limits, but also incorporating an early stopping mechanism. Additionally, human intervention can refine the optimization process by examining the results and limiting the search space, which are functions supported by our AutoML framework. Combining computational resources with human-guided intervention provides a path to achieving the best performance in ML tasks

Our AutoML is designed to be user-friendly, allowing users with little to no prior knowledge to get started effortlessly using just 4-5 lines of code. It also offers flexibility for customization to cater to personalized use scenarios, accommodating both novices and advanced users with varying needs. Furthermore, beyond its immediate application in ML solution development, our AutoML can serve as a benchmark for evaluating future innovations in ML models within the insurance domain. By providing a consistent and comprehensive evaluation platform, it has the potential to serve as a valuable reference point for assessing the performance and advancements of emerging algorithms.

\clearpage

\bibliographystyle{apalike}
\bibliography{reference.bib}

\begin{thebibliography}{}

\bibitem[Akiba et~al., 2019]{Akiba2019}
Akiba, T., Sano, S., Yanase, T., Ohta, T., and Koyama, M. (2019).
\newblock Optuna: A next-generation hyperparameter optimization framework.
\newblock In {\em Proceedings of the 25th ACM SIGKDD International Conference on Knowledge Discovery \& Data Mining}, Proceedings of the International Conference on Knowledge Discovery \& Data Mining, pages 2623--2631. Association for Computing Machinery.

\bibitem[Azur et~al., 2011]{Azur2011}
Azur, M.~J., Stuart, E.~A., Frangakis, C., and Leaf, P.~J. (2011).
\newblock Multiple imputation by chained equations: what is it and how does it work?
\newblock {\em International Journal of Methods in Psychiatric Research}, 20:40--49.

\bibitem[Bakhteev and Strijov, 2020]{bakhteevComprehensiveAnalysisGradientbased2020}
Bakhteev, O.~Y. and Strijov, V.~V. (2020).
\newblock Comprehensive analysis of gradient-based hyperparameter optimization algorithms.
\newblock {\em Annals of Operations Research}, 289(1):51--65.

\bibitem[Bams et~al., 2009]{bamsLossFunctionsOption2009}
Bams, D., Lehnert, T., and Wolff, C. C.~P. (2009).
\newblock Loss {{Functions}} in {{Option Valuation}}: {{A Framework}} for {{Selection}}.
\newblock {\em Management Science}, 55(5):853--862.

\bibitem[Batista et~al., 2004]{Batista2004}
Batista, G. E. A. P.~A., Prati, R.~C., and Monard, M.~C. (2004).
\newblock A study of the behavior of several methods for balancing machine learning training data.
\newblock {\em ACM SIGKDD Explorations Newsletter}, 6:20--29.

\bibitem[Bergstra et~al., 2013]{Bergstra2013}
Bergstra, J., Yamins, D., and Cox, D. (2013).
\newblock Making a science of model search: Hyperparameter optimization in hundreds of dimensions for vision architectures.
\newblock In Dasgupta, S. and McAllester, D., editors, {\em Proceedings of the 30th International Conference on Machine Learning}, volume~28 of {\em Proceedings of Machine Learning Research}, pages 115--123. PMLR.

\bibitem[Chandrashekar and Sahin, 2014]{Chandrashekar2014}
Chandrashekar, G. and Sahin, F. (2014).
\newblock A survey on feature selection methods.
\newblock {\em Computers \& Electrical Engineering}, 40:16--28.

\bibitem[Charpentier, 2015]{charpentier2014computational}
Charpentier, A. (2015).
\newblock Computational actuarial science with r.
\newblock {\em Journal of the Royal Statistical Society Series A: Statistics in Society}, 178:782--783.

\bibitem[Charpentier et~al., 2023]{charpentierReinforcementLearningEconomics2023}
Charpentier, A., {\'E}lie, R., and Remlinger, C. (2023).
\newblock Reinforcement {{Learning}} in {{Economics}} and {{Finance}}.
\newblock {\em Computational Economics}, 62(1):425--462.

\bibitem[Chawla et~al., 2002]{Chawla2002}
Chawla, N.~V., Bowyer, K.~W., Hall, L.~O., and Kegelmeyer, W.~P. (2002).
\newblock Smote: Synthetic minority over-sampling technique.
\newblock {\em Journal of Artificial Intelligence Research}, 16:321--357.

\bibitem[Chen and Guestrin, 2016]{Chen2016}
Chen, T. and Guestrin, C. (2016).
\newblock Xgboost: A scalable tree boosting system.
\newblock In {\em Proceedings of the 22nd ACM SIGKDD International Conference on Knowledge Discovery and Data Mining}, pages 785--794. ACM.

\bibitem[Cummings and Hartman, 2022]{cummingsUsingMachineLearning2022}
Cummings, J. and Hartman, B. (2022).
\newblock Using {{Machine Learning}} to {{Better Model Long-Term Care Insurance Claims}}.
\newblock {\em North American Actuarial Journal}, 26(3):470--483.

\bibitem[De~Jong et~al., 2008]{de2008generalized}
De~Jong, P., Heller, G.~Z., et~al. (2008).
\newblock {\em Generalized linear models for insurance data}.
\newblock Cambridge University Press.

\bibitem[Dong et~al., 2020]{dong2020survey}
Dong, X., Yu, Z., Cao, W., Shi, Y., and Ma, Q. (2020).
\newblock A survey on ensemble learning.
\newblock {\em Frontiers of Computer Science}, 14:241--258.

\bibitem[Feurer et~al., 2022]{feurerAutosklearnHandsfreeAutoML2022}
Feurer, M., Eggensperger, K., Falkner, S., Lindauer, M., and Hutter, F. (2022).
\newblock Auto-sklearn 2.0: Hands-free {{AutoML}} via meta-learning.
\newblock {\em The Journal of Machine Learning Research}, 23(1):261:11936--261:11996.

\bibitem[Frees et~al., 2016]{frees2016multivariate}
Frees, E.~W., Lee, G., and Yang, L. (2016).
\newblock Multivariate frequency-severity regression models in insurance.
\newblock {\em Risks}, 4(1):4.

\bibitem[Galar et~al., 2012]{galar2011review}
Galar, M., Fernandez, A., Barrenechea, E., Bustince, H., and Herrera, F. (2012).
\newblock A review on ensembles for the class imbalance problem: bagging-, boosting-, and hybrid-based approaches.
\newblock {\em IEEE Transactions on Systems, Man, and Cybernetics, Part C (Applications and Reviews)}, 42(4):463--484.

\bibitem[Gan and Valdez, 2024]{guojunCompositionalDataRegression2024}
Gan, G. and Valdez, E.~A. (2024).
\newblock Compositional {{Data Regression}} in {{Insurance}} with {{Exponential Family PCA}}.
\newblock {\em Variance}, 17(1).

\bibitem[Garc{\'\i}a et~al., 2016]{garcia2016big}
Garc{\'\i}a, S., Ram{\'\i}rez-Gallego, S., Luengo, J., Ben{\'\i}tez, J.~M., and Herrera, F. (2016).
\newblock Big data preprocessing: methods and prospects.
\newblock {\em Big Data Analytics}, 1(1):1--22.

\bibitem[Guerra and Castelli, 2021]{guerraMachineLearningApplied2021}
Guerra, P. and Castelli, M. (2021).
\newblock Machine {{Learning Applied}} to {{Banking Supervision}} a {{Literature Review}}.
\newblock {\em Risks}, 9(7):136.

\bibitem[Guo et~al., 2017]{Haixiang2017}
Guo, H., Li, Y., Shang, J., Gu, M., Huang, Y., and Gong, B. (2017).
\newblock Learning from class-imbalanced data: Review of methods and applications.
\newblock {\em Expert Systems with Applications}, 73:220--239.

\bibitem[Hart, 1968]{Hart1968}
Hart, P.~E. (1968).
\newblock The condensed nearest neighbor rule.
\newblock {\em IEEE Transactions on Information Theory}, 14:515--516.

\bibitem[Hartman et~al., 2020]{hartmanPredictingHighCostHealth2020}
Hartman, B., Owen, R., and Gibbs, Z. (2020).
\newblock Predicting {{High-Cost Health Insurance Members}} through {{Boosted Trees}} and {{Oversampling}}: {{An Application Using}} the {{HCCI Database}}.
\newblock {\em North American Actuarial Journal}, 25(1):53--61.

\bibitem[He and Garcia, 2009]{He2009}
He, H. and Garcia, E.~A. (2009).
\newblock Learning from imbalanced data.
\newblock {\em IEEE Transactions on Knowledge and Data Engineering}, 21:1263--1284.

\bibitem[He et~al., 2021]{HE2021106622}
He, X., Zhao, K., and Chu, X. (2021).
\newblock Automl: A survey of the state-of-the-art.
\newblock {\em Knowledge-Based Systems}, 212:106622.

\bibitem[Hodge and Austin, 2004]{hodge2004survey}
Hodge, V. and Austin, J. (2004).
\newblock A survey of outlier detection methodologies.
\newblock {\em Artificial intelligence review}, 22:85--126.

\bibitem[Hu et~al., 2022]{huImbalancedLearningInsurance2022a}
Hu, C., Quan, Z., and Chong, W.~F. (2022).
\newblock Imbalanced learning for insurance using modified loss functions in tree-based models.
\newblock {\em Insurance: Mathematics and Economics}, 106:13--32.

\bibitem[Jordan and Mitchell, 2015]{jordan2015machine}
Jordan, M.~I. and Mitchell, T.~M. (2015).
\newblock Machine learning: Trends, perspectives, and prospects.
\newblock {\em Science}, 349(6245):255--260.

\bibitem[Ke et~al., 2017]{Ke2017}
Ke, G., Meng, Q., Finley, T., Wang, T., Chen, W., Ma, W., Ye, Q., and Liu, T.-Y. (2017).
\newblock Lightgbm: A highly efficient gradient boosting decision tree.
\newblock In {\em Advances in Neural Information Processing Systems}, volume~30.

\bibitem[Kononenko, 2001]{kononenko2001machine}
Kononenko, I. (2001).
\newblock Machine learning for medical diagnosis: history, state of the art and perspective.
\newblock {\em Artificial Intelligence in medicine}, 23(1):89--109.

\bibitem[LeDell and Poirier, 2020]{H2OAutoML20}
LeDell, E. and Poirier, S. (2020).
\newblock {H2O} {A}uto{ML}: Scalable automatic machine learning.
\newblock {\em 7th ICML Workshop on Automated Machine Learning (AutoML)}.

\bibitem[Liaw et~al., 2018]{Liaw2018}
Liaw, R., Liang, E., Nishihara, R., Moritz, P., Gonzalez, J.~E., and Stoica, I. (2018).
\newblock Tune: A research platform for distributed model selection and training.
\newblock {\em arXiv preprint arXiv:1807.05118}.

\bibitem[Lin et~al., 2017]{linFocalLossDense2017}
Lin, T.-Y., Goyal, P., Girshick, R., He, K., and Dollar, P. (2017).
\newblock Focal {{Loss}} for {{Dense Object Detection}}.
\newblock In {\em Proceedings of the {{IEEE International Conference}} on {{Computer Vision}}}, pages 2980--2988.

\bibitem[Ma and Sun, 2020]{maMachineLearningAI2020}
Ma, L. and Sun, B. (2020).
\newblock Machine learning and {{AI}} in marketing {\textendash} {{Connecting}} computing power to human insights.
\newblock {\em International Journal of Research in Marketing}, 37(3):481--504.

\bibitem[Masello et~al., 2023]{maselloUsingContextualData2023}
Masello, L., Castignani, G., Sheehan, B., Guillen, M., and Murphy, F. (2023).
\newblock Using contextual data to predict risky driving events: {{A}} novel methodology from explainable artificial intelligence.
\newblock {\em Accident Analysis \& Prevention}, 184:106997.

\bibitem[Mitchell et~al., 1990]{Mitchell1990}
Mitchell, T., Buchanan, B., Dejong, G., Dietterich, T., Rosenbloom, P., and Waibel, A. (1990).
\newblock Machine learning.
\newblock {\em Annual Review of Computer Science}, 4:417--433.

\bibitem[Noll et~al., 2020]{noll2020case}
Noll, A., Salzmann, R., and Wuthrich, M.~V. (2020).
\newblock Case study: French motor third-party liability claims.
\newblock {\em Available at SSRN: https://ssrn.com/abstract=3164764 or http://dx.doi.org/10.2139/ssrn.3164764}.

\bibitem[Paszke et~al., 2019]{NEURIPS2019_9015}
Paszke, A., Gross, S., Massa, F., Lerer, A., Bradbury, J., Chanan, G., Killeen, T., Lin, Z., Gimelshein, N., Antiga, L., Desmaison, A., Kopf, A., Yang, E., DeVito, Z., Raison, M., Tejani, A., Chilamkurthy, S., Steiner, B., Fang, L., Bai, J., and Chintala, S. (2019).
\newblock Pytorch: An imperative style, high-performance deep learning library.
\newblock In Wallach, H., Larochelle, H., Beygelzimer, A., d\textquotesingle Alch\'{e}-Buc, F., Fox, E., and Garnett, R., editors, {\em Advances in Neural Information Processing Systems 32}, pages 8026--8037. Curran Associates, Inc.

\bibitem[Pedregosa et~al., 2011]{sklearn}
Pedregosa, F., Michel, V., Grisel, O., Blondel, M., Prettenhofer, P., Weiss, R., Vanderplas, J., Cournapeau, D., Pedregosa, F., Varoquaux, G., Gramfort, A., Thirion, B., Grisel, O., Dubourg, V., Passos, A., Brucher, M., Perrot, M., and Édouard Duchesnay (2011).
\newblock Scikit-learn: Machine learning in python.
\newblock {\em Journal of Machine Learning Research}, 12:2825--2830.

\bibitem[Peiris et~al., 2024]{peirisIntegrationTraditionalTelematics2024}
Peiris, H., Jeong, H., Kim, J.-K., and Lee, H. (2024).
\newblock Integration of traditional and telematics data for efficient insurance claims prediction.
\newblock {\em ASTIN Bulletin: The Journal of the IAA}, 54(2):263--279.

\bibitem[Polikar, 2006]{polikar2006ensemble}
Polikar, R. (2006).
\newblock Ensemble based systems in decision making.
\newblock {\em IEEE Circuits and systems magazine}, 6(3):21--45.

\bibitem[Qayyum et~al., 2020]{qayyum2020secure}
Qayyum, A., Qadir, J., Bilal, M., and Al-Fuqaha, A. (2020).
\newblock Secure and robust machine learning for healthcare: A survey.
\newblock {\em IEEE Reviews in Biomedical Engineering}, 14:156--180.

\bibitem[Quan and Valdez, 2018]{quan2018predictive}
Quan, Z. and Valdez, E.~A. (2018).
\newblock Predictive analytics of insurance claims using multivariate decision trees.
\newblock {\em Dependence Modeling}, 6(1):377--407.

\bibitem[Quan et~al., 2023]{quanHybridTreebasedMethods2023}
Quan, Z., Wang, Z., Gan, G., and Valdez, E.~A. (2023).
\newblock On hybrid tree-based methods for short-term insurance claims.
\newblock {\em Probability in the Engineering and Informational Sciences}, 37(2):597--620.

\bibitem[Rapin and Teytaud, 2018]{nevergrad}
Rapin, J. and Teytaud, O. (2018).
\newblock {Nevergrad - A gradient-free optimization platform}.
\newblock \url{https://GitHub.com/FacebookResearch/Nevergrad}.

\bibitem[Sagi and Rokach, 2018]{Sagi2018}
Sagi, O. and Rokach, L. (2018).
\newblock Ensemble learning: A survey.
\newblock {\em Wiley Interdisciplinary Reviews: Data Mining and Knowledge Discovery}, 8:e1249.

\bibitem[Salehi et~al., 2017]{salehiTverskyLossFunction2017}
Salehi, S. S.~M., Erdogmus, D., and Gholipour, A. (2017).
\newblock Tversky {{Loss Function}} for {{Image Segmentation Using 3D Fully Convolutional Deep Networks}}.
\newblock In Wang, Q., Shi, Y., Suk, H.-I., and Suzuki, K., editors, {\em Machine {{Learning}} in {{Medical Imaging}}}, pages 379--387. Springer International Publishing.

\bibitem[Servén and Brummitt, 2018]{pygam}
Servén, D. and Brummitt, C. (2018).
\newblock pygam: Generalized additive models in python.

\bibitem[Shi et~al., 2024]{shiLeveragingWeatherDynamics2024b}
Shi, P., Zhang, W., and Shi, K. (2024).
\newblock Leveraging {{Weather Dynamics}} in {{Insurance Claims Triage Using Deep Learning}}.
\newblock {\em Journal of the American Statistical Association}, 119(546):825--838.

\bibitem[Si et~al., 2022]{siAutomobileInsuranceClaim2022}
Si, J., He, H., Zhang, J., and Cao, X. (2022).
\newblock Automobile insurance claim occurrence prediction model based on ensemble learning.
\newblock {\em Applied Stochastic Models in Business and Industry}, 38(6):1099--1112.

\bibitem[Snoek et~al., 2012]{NIPS2012_05311655}
Snoek, J., Larochelle, H., and Adams, R.~P. (2012).
\newblock Practical bayesian optimization of machine learning algorithms.
\newblock In Pereira, F., Burges, C., Bottou, L., and Weinberger, K., editors, {\em Advances in Neural Information Processing Systems}, volume~25. Curran Associates, Inc.

\bibitem[So, 2024]{soEnhancedGradientBoosting}
So, B. (2024).
\newblock Enhanced gradient boosting for zero-inflated insurance claims and comparative analysis of {{CatBoost}}, {{XGBoost}}, and {{LightGBM}}.
\newblock {\em Scandinavian Actuarial Journal}, pages 1--23.

\bibitem[So et~al., 2021]{soCOSTSENSITIVEMULTICLASSADABOOST2021a}
So, B., Boucher, J.-P., and Valdez, E.~A. (2021).
\newblock Cost-sensitive multi-class adaboost for understanding driving behavior based on telematics.
\newblock {\em ASTIN Bulletin: The Journal of the IAA}, 51(3):719--751.

\bibitem[So and Valdez, 2024]{soSAMMEC2Algorithm2024}
So, B. and Valdez, E.~A. (2024).
\newblock {{SAMME}}.{{C2}} algorithm for imbalanced multi-class classification.
\newblock {\em Soft Computing}.

\bibitem[Stekhoven and Bühlmann, 2012]{Stekhoven2012}
Stekhoven, D.~J. and Bühlmann, P. (2012).
\newblock Missforest—non-parametric missing value imputation for mixed-type data.
\newblock {\em Bioinformatics}, 28:112--118.

\bibitem[Thornton et~al., 2013]{Thornton2013}
Thornton, C., Hutter, F., Hoos, H.~H., and Leyton-Brown, K. (2013).
\newblock Auto-weka: Combined selection and hyperparameter optimization of classification algorithms.
\newblock In {\em Proceedings of the 19th ACM SIGKDD international conference on Knowledge discovery and data mining}, pages 847--855. Association for Computing Machinery.

\bibitem[Tomek, 1976]{TOMEK1976}
Tomek, I. (1976).
\newblock Two modifications of cnn.
\newblock {\em IEEE Transactions on Systems, Man, and Cybernetics}, SMC-6:769 -- 772.

\bibitem[Turcotte and Boucher, 2024]{turcotteGAMLSSLongitudinalMultivariate2024}
Turcotte, R. and Boucher, J.-P. (2024).
\newblock {{GAMLSS}} for {{Longitudinal Multivariate Claim Count Models}}.
\newblock {\em North American Actuarial Journal}, 28(2):337--360.

\bibitem[Wang et~al., 2022]{wangComprehensiveSurveyLoss2022}
Wang, Q., Ma, Y., Zhao, K., and Tian, Y. (2022).
\newblock A {{Comprehensive Survey}} of {{Loss Functions}} in {{Machine Learning}}.
\newblock {\em Annals of Data Science}, 9(2):187--212.

\bibitem[Wilson, 1972]{Wilson1972}
Wilson, D.~L. (1972).
\newblock Asymptotic properties of nearest neighbor rules using edited data.
\newblock {\em IEEE Transactions on Systems, Man and Cybernetics}, SMC-2:408--421.

\bibitem[Wu et~al., 2019]{wu2019hyperparameter}
Wu, J., Chen, X.-Y., Zhang, H., Xiong, L.-D., Lei, H., and Deng, S.-H. (2019).
\newblock Hyperparameter optimization for machine learning models based on bayesian optimization.
\newblock {\em Journal of Electronic Science and Technology}, 17(1):26--40.

\bibitem[W{\"u}thrich, 2019]{wuthrich2019generalized}
W{\"u}thrich, M.~V. (2019).
\newblock From generalized linear models to neural networks, and back.
\newblock Technical report, Department of Mathematics, ETH Zurich.

\bibitem[Yang and Shami, 2020]{yang2020hyperparameter}
Yang, L. and Shami, A. (2020).
\newblock On hyperparameter optimization of machine learning algorithms: Theory and practice.
\newblock {\em Neurocomputing}, 415:295--316.

\bibitem[Yoon et~al., 2018]{Yoon2018}
Yoon, J., Jordon, J., and Schaar, M. V.~D. (2018).
\newblock Gain: Missing data imputation using generative adversarial nets.
\newblock In {\em Proceedings of the 35th International Conference on Machine Learning}, volume~80, pages 5689--5698. PMLR.

\bibitem[Young et~al., 2015]{youngOptimizingDeepLearning2015}
Young, S.~R., Rose, D.~C., Karnowski, T.~P., Lim, S.-H., and Patton, R.~M. (2015).
\newblock Optimizing deep learning hyper-parameters through an evolutionary algorithm.
\newblock In {\em Proceedings of the {{Workshop}} on {{Machine Learning}} in {{High-Performance Computing Environments}}}, {{MLHPC}} '15, pages 1--5, {New York, NY, USA}. {Association for Computing Machinery}.

\bibitem[Zhang et~al., 2024]{zhangBayesianCARTModels2024}
Zhang, Y., Ji, L., Aivaliotis, G., and Taylor, C. (2024).
\newblock Bayesian {{CART}} models for insurance claims frequency.
\newblock {\em Insurance: Mathematics and Economics}, 114:108--131.

\bibitem[Z{\"o}ller and Huber, 2021]{zoller2021benchmark}
Z{\"o}ller, M.-A. and Huber, M.~F. (2021).
\newblock Benchmark and survey of automated machine learning frameworks.
\newblock {\em Journal of artificial intelligence research}, 70:409--472.

\end{thebibliography}

\clearpage

\begin{appendices}
\section{Notations}\label{appendix_sec:notation}
\begin{table}[!ht]
\small
\centering
\begin{tabular}{c c } 
\hline
Notation & Description  \\
\hline
$\mathcal{C}$ & a conjunction space of model space and hyperparameter space \\[-0.1ex]
$\mathcal{D}$ & a dataset \\[-0.1ex]
$\mathcal{L}$ & a loss function \\[-0.1ex]
$\mathcal{M}$ & the model space for all ML models \\[-0.1ex]
$\mathcal{P}$ & a ML workflow pipeline \\[-0.1ex]
$\mathcal{V}$ & an objective function \\[-0.1ex]
$\mathcal{U}$ & the search space \\[-0.1ex]
$A$ & the majority class in the context of an imbalanced problem \\[-0.1ex]
$B$ & a data balancing algorithm \\[-0.1ex]
$E$ & a data encoding algorithm \\[-0.1ex]
$F$ & a feature selection algorithm \\[-0.1ex]
$G$ & evaluation budget \\[-0.1ex]
$H$ & the number of pipelines to construct the ensemble model \\[-0.1ex]
$I$ & an imputation algorithm \\[-0.1ex]
$M$ & an ML model \\[-0.1ex]
$N$ & the number of tunable hyperparameters \\[-0.1ex]
$O$ & unique categories of a classification task \\[-0.1ex]
$\textbf{Q}$ & a feature selection rule \\[-0.1ex]
$R$ & the imbalance threshold \\[-0.1ex]
$S$ & a data scaling algorithm \\[-0.1ex]
$T$ & time budget \\[-0.1ex]
$W$ & the number of features/columns of a feature matrix $\textbf{X}$ \\[-0.1ex]
$(\textbf{X}, \textbf{y})$ & a pair of feature matrix $\textbf{X}$ and response vector $\textbf{y}$ in a dataset \\[-0.1ex]
$Z$ & the number of observations/rows of a feature matrix \\[-0.1ex]
$\gamma$ & a voting mechanism \\[-0.1ex]
$\theta$ & a parameter set of the ML model \\[-0.1ex]
$\lambda$ &  a hyperparameter set of ML model \\[-0.1ex]
$\Lambda_{j}$ & range of hyperparameter $j$ \\[-0.1ex]
$\bm{\Lambda}$ & the space defined by the ranges of all hyperparameters \\[-0.1ex]
$\Sigma$ & a pipeline ensemble model \\[-0.1ex]
$\Omega$ & the number of features selected by a feature selection rule \\[-0.1ex]
\hline
\end{tabular}
\caption{Notations used in the formulation of AutoML}
\label{tab:notation}
\end{table}

\clearpage

\section{Three ensemble strategies}\label{appendix_sec:enc}

The following paragraphs elaborate on how the training of the three ensemble structures and the voting mechanism are adopted in our work:

\begin{figure}[htbp]
\centering
\includegraphics[width=\linewidth]{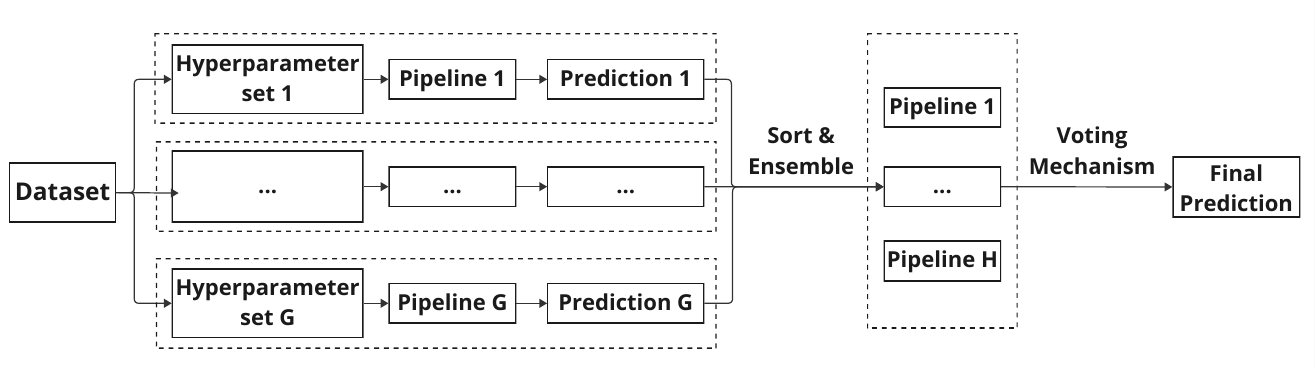}
\caption{An illustration of stacking ensemble training diagram}
\label{fig:stacking}
\end{figure}

\textbf{Stacking Ensemble}: As illustrated in Figure \ref{fig:stacking}, the stacking ensemble represents the most intuitive ensemble structure, where individual pipelines are trained independently on the original full datasets. The only difference distinguishing different individual pipelines is the hyperparameter set sampled by the search algorithm, which results in varying evaluation losses. Consequently, all pipelines can be trained in parallel, leveraging the multi-core, multi-threading capabilities of modern computing systems. 

Once the time or evaluation budget is exhausted, a total of $G$ pipelines, denoted as $\{\mathcal{P}_{g}\}$, $g=1, 2, ..., G$, have been trained. These pipelines are then ranked based on their evaluation losses, producing a sorted list $\{\mathcal{P}_{(h)}\}$, $h=1, 2, ...$. Given our AutoML’s optimization direction, the sorted pipelines $\{\mathcal{P}_{(h)}\}$ are arranged in ascending order of $h$, indicating that $\mathcal{P}_{(1)}$ has the lowest evaluation loss and performs the best. The ensemble model, denoted as $\Sigma_{H}$, is constructed utilizing the best-performing $H$ pipelines, $\{\mathcal{P}_{(1)}, \mathcal{P}_{(2)}, ..., \mathcal{P}_{(H)}\}$. The assembly of the ensemble model $\Sigma_{H}$ involves no additional computation but merely the stacking the selected pipelines, which can be expressed as
$$
 \Sigma_{H}=\Sigma_{H}(\mathcal{P}_{(1)}, \mathcal{P}_{(2)}, \ldots, \mathcal{P}_{(H)})
$$
Algorithm \ref{alg:stacking} summarizes the stacking ensemble process, highlighting that the key differences from Algorithm \ref{alg:automl} are the post-ranking step and the stacking procedure for the ensemble model, specifically line 14 in the algorithm description.

{\SetAlgoNoLine
\begin{algorithm}
\caption{The Stacking Ensemble}\label{alg:stacking}
\KwIn{Dataset $\mathcal{D}=(\mathcal{D}_{train}, \mathcal{D}_{valid})$; Search space $\mathcal{U}$; Time budget $T$; Evaluation budget $G$; Search algorithm $Samp$; Size of the ensemble model $H$}
\KwOut{Ensemble $\Sigma_{H}$}
$k = 0$ \Comment*[r]{Round of evaluation}
$t^{re} = T$ \Comment*[r]{Remaining time budget}
$g^{re} = G$ \Comment*[r]{Remaining evaluation budget}
\While{$t^{re} > 0$ and $g^{re} > 0$}{
    $t^{start}=CurrentTime$;\\
    $(E^{(k)}, \lambda_{E}^{(k)}), (I^{(k)}, \lambda_{I}^{(k)}), (B^{(k)}, \lambda_{B}^{(k)}), (S^{(k)}, \lambda_{S}^{(k)}), (F^{(k)}, \lambda_{F}^{(k)}), (M^{(k)}, \lambda_{M}^{(k)})=Samp^{(k)}(\mathcal{U})$;\\
    $\mathcal{P}_{k}=M^{(k)}_{\lambda_{M}^{(k)}}\circ F^{(k)}_{\lambda_{F}^{(k)}}\circ S^{(k)}_{\lambda_{S}^{(k)}}\circ B^{(k)}_{\lambda_{B}^{(k)}}\circ I^{(k)}_{\lambda_{I}^{(k)}}\circ E^{(k)}_{\lambda_{E}^{(k)}}$;\\
    $L^{eval, (k)}=\mathcal{V}(\mathcal{L}, \mathcal{P}_{k}, \mathcal{D})$;\\
    $t^{end}=CurrentTime$;\\
    $k = k + 1$;\\
    $t^{re} = t^{re} - (t^{end} - t^{start})$;\\
    $g^{re} = g^{re} - 1$;\\
}
$\{\mathcal{P}_{(k)}\}=sort(\{\mathcal{P}_{k}\})$;\\
$\Sigma_{H}=\Sigma_{H}(\mathcal{P}_{(1)}, \mathcal{P}_{(2)}, \ldots, \mathcal{P}_{(H)})$;\\
\Return{$\Sigma_{H}$};\\
\end{algorithm}}

\begin{figure}[htbp]
\centering
\includegraphics[width=\linewidth]{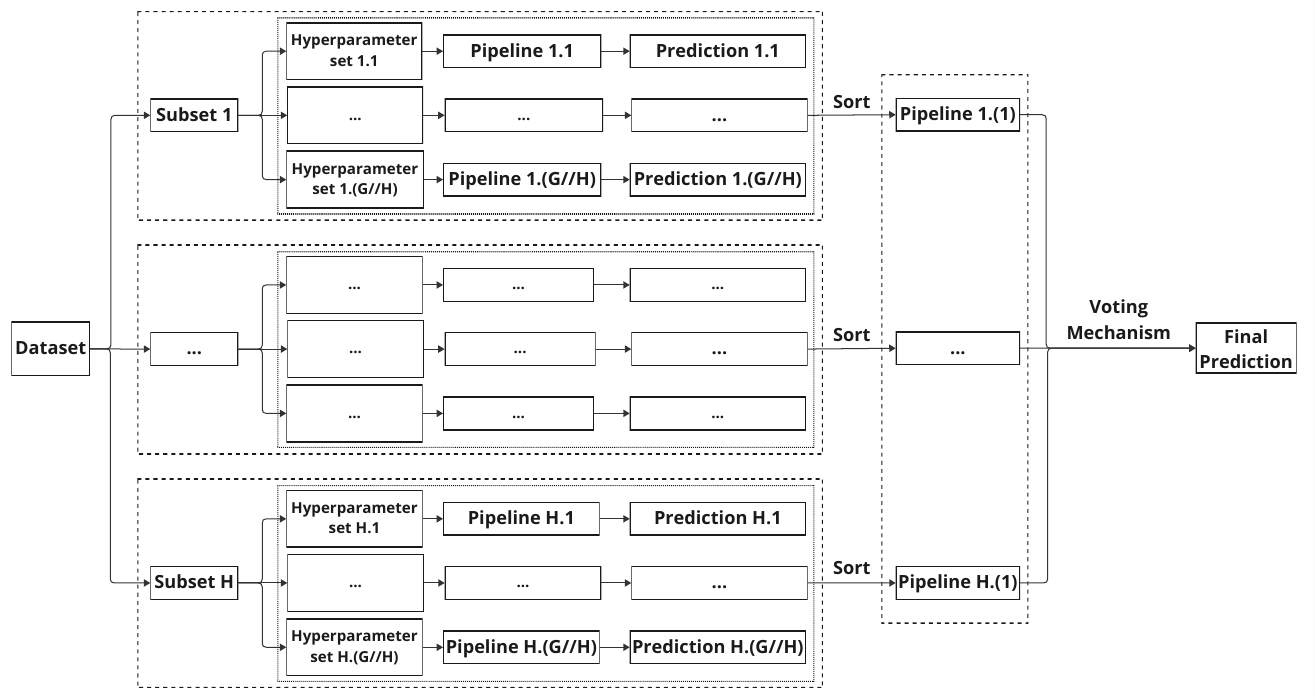}
\caption{An illustration of bagging ensemble training diagram}
\label{fig:bagging}
\end{figure}

\textbf{Bagging Ensemble}: The bagging ensemble follows a similar training diagram as the stacking ensemble, but differs by optimizing the individual pipelines on the subsets of the features, rather than the full feature set. As demonstrated in Figure \ref{fig:bagging}, for a pre-defined number of $H$, denoting the number of pipelines in the ensemble model $\Sigma_{H}$, $H$ subsets of datasets are constructed by randomly selecting only a proportion of the features. Each pipeline $\mathcal{P}_{h,l}$, $h=1,2,\ldots,H$ and $l=1,2,\ldots, G//H$, is optimized on the subset dataset $\mathcal{D}^{(h)}$ which is constructed from the original dataset $\mathcal{D}=(\textbf{X}, \textbf{y})$ as follows: for every subset, a feature selection rule $\textbf{Q}^{(h)}$, in the form of $W\times1$ vector where $W$ denotes the number of features in the feature matrix $\textbf{X}$, is defined by our AutoML. The feature selection rules consist of binary values where the $w$-th element of rule $q^{(h)}_{w}=1$ if the $w$-th feature is selected by the rule and $q^{(h)}_{w}=0$ otherwise. Following the feature selection rule $\textbf{Q}^{(h)}$, a subset matrix $\bm{\rho}^{(h)}$ of shape of $W\times \Omega^{(h)}$ can be defined, where elements $\rho_{w, \sum_{\alpha=1}^{w}q^{(h)}_{\alpha}}^{(h)}=q^{(h)}_{w}$ for $w=1, 2, ..., W$ and $\rho_{w, \omega}=0$ for all $(w, \omega)\notin\{(w, \sum_{\alpha=1}^{w}q^{(h)}_{\alpha})\}_{w}$ ($w=1, 2, ..., W$ and $\omega=1, 2, ..., \Omega^{(h)}$). $\Omega^{(h)}$ is the number of features selected by rule $\textbf{Q}^{(h)}$, so $\Omega^{(h)}=\sum_{\alpha=1}^{w}q^{(h)}_{\alpha}$. The subset dataset is then defined as $\mathcal{D}^{(h)}=(\textbf{X}\bm{\rho}^{(h)}, \textbf{y})$. 

Given the structure of tabular datasets, each pipeline aims to be optimized on the horizontal partition of the original datasets instead of the full original datasets, as seen in the stacking ensemble. The pipelines of each subset, once trained, are ordered by their evaluation loss within the groups of subsets, resulting in ${\mathcal{P}_{h, (l)}}$, $h=1,2,\ldots,H$ and $l=1,2,\ldots, G//H$. Using the same ascending order, each pipeline ${\mathcal{P}_{h, (1)}}$ is the best-performing pipeline for subset dataset $\mathcal{D}^{(h)}$. The ensemble model, targeting optimal performance, is then constructed as:
$$
\Sigma_{H}=\Sigma_{H}(\mathcal{P}_{1, (1)}, \mathcal{P}_{2, (1)}, \ldots, \mathcal{P}_{H, (1)})
$$
by selecting the best-performing pipelines from each subset. The construction of the ensemble model becomes straightforward by stacking the best-performing pipelines from each subset dataset, once the optimization and ranking on subsets are completed. The bagging algorithm is summarized in Algorithm \ref{alg:bagging}. Unlike Algorithm \ref{alg:stacking}, additional subset matrices $\{{\bm{\rho}^{(h)}}\}$ ($h=1, 2, ..., H$) are required to construct the subset datasets, and the optimization loops are performed $H$ times to obtain the pipelines $\{\mathcal{P}_{h, (1)}\}$. Bagging is typically used to prevent over-fitting, as optimizing on data subsets is sub-optimal compared to utilizing the full datasets. However, as demonstrated in \citet{galar2011review}, bagging algorithms are also beneficial for imbalance learning.

{\SetAlgoNoLine
\begin{algorithm}
\caption{The Bagging Ensemble}\label{alg:bagging}
\KwIn{Dataset $\mathcal{D}=(\mathcal{D}_{train}, \mathcal{D}_{valid})$; Search space $\mathcal{U}$; Time budget $T$; Evaluation budget $G$; Search algorithm $Samp$; Size of the ensemble $H$; Subset Matrices $\{\bm{\rho}^{(h)}\}_{h=1,2,\ldots,H}$}
\KwOut{Ensemble $\Sigma_{H}$}
\For{$h \gets 1$ to $H$}{
$k = 0$ \Comment*[r]{Round of evaluation}
$t^{re} = T//H$ \Comment*[r]{Remaining time budget}
$g^{re} = G//H$ \Comment*[r]{Remaining evaluation budget}
$\mathcal{D}^{(h)}=((\textbf{X}_{train}\bm{\rho}^{(h)}, \textbf{y}_{train}),(\textbf{X}_{valid}\bm{\rho}^{(h)}, \textbf{y}_{valid}))$;\\
\While{$t^{re} > 0$ and $g^{re} > 0$}{
    $t^{start}=CurrentTime$;\\
    $(E^{(k)}, \lambda_{E}^{(k)}), (I^{(k)}, \lambda_{I}^{(k)}), (B^{(k)}, \lambda_{B}^{(k)}), (S^{(k)}, \lambda_{S}^{(k)}), (F^{(k)}, \lambda_{F}^{(k)}), (M^{(k)}, \lambda_{M}^{(k)})=Samp^{(k)}(\mathcal{U})$;\\
    $\mathcal{P}_{h, k}=M^{(k)}_{\lambda_{M}^{(k)}}\circ F^{(k)}_{\lambda_{F}^{(k)}}\circ S^{(k)}_{\lambda_{S}^{(k)}}\circ B^{(k)}_{\lambda_{B}^{(k)}}\circ I^{(k)}_{\lambda_{I}^{(k)}}\circ E^{(k)}_{\lambda_{E}^{(k)}}$;\\
    $L^{eval, (k)}=\mathcal{V}(\mathcal{L}, \mathcal{P}_{h, k}, \mathcal{D}^{(h)})$;\\
    $t^{end}=CurrentTime$;\\
    $k = k + 1$;\\
    $t^{re} = t^{re} - (t^{end} - t^{start})$;\\
    $g^{re} = g^{re} - 1$;\\
}
$\{\mathcal{P}_{h, (k)}\}=sort(\{\mathcal{P}_{h, k}\})$;\\
}
$\Sigma_{H}=\Sigma_{H}(\mathcal{P}_{1, (1)}, \mathcal{P}_{2, (1)}, \ldots, \mathcal{P}_{H, (1)})$;\\
\Return{$\Sigma_{H}$};\\
\end{algorithm}}

\begin{figure}[htbp]
\centering
\includegraphics[width=\linewidth]{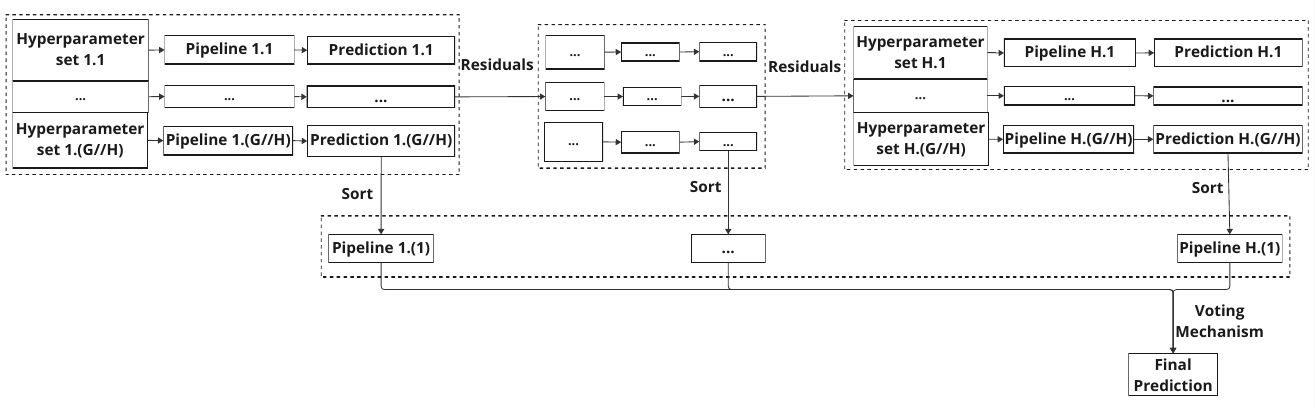}
\caption{An illustration of boosting ensemble training diagram}
\label{fig:boosting}
\end{figure}

\textbf{Boosting Ensemble}: In contrast to the parallel training diagram utilized in the stacking and bagging ensembles, boosting ensemble, as shown in Figure \ref{fig:boosting}, employs a sequential learning approach where each subsequent pipeline is optimized based on the residuals/gradients of the previous pipelines. The sequential design of boosting ensemble is motivated by the idea that later models are trained to correct the errors of earlier models, addressing issues that earlier models may not have effectively resolved. For a boosting ensemble of size $H$, the pipeline $\mathcal{P}_{h}$, ($h=1,2,\ldots,H$), at step $h$, does not train on the original training set $\mathcal{D}_{train}=(\textbf{X}_{train}, \textbf{y}_{train})$, but instead focuses on optimizing the residual dataset.
$$
\mathcal{D}_{h}=(\textbf{X}_{train}, \textbf{y}_{train, h})
$$
where 
$$
\textbf{y}_{train, h}=\textbf{y}_{train, h-1} - \mathcal{P}_{h-1}(\textbf{X}_{train})
$$ 
for $h = 1, 2, ..., H$. The initialization of the residuals and predictions can be defined as $\textbf{y}_{train, 0}=\textbf{y}_{train}$ and $\mathcal{P}_{0}(\textbf{X}_{train})=0$, indicating the first pipeline is trained on the original train set $\mathcal{D}_{train}$. By learning from residuals instead of original response variables, the subsequent pipelines possess the potential to address errors that preceding pipelines could not, thereby potentially solving the problem of imbalance. In our AutoML, rather than utilizing a full sequential training diagram, we split the training by steps. At step $h$, ($h=1, 2, ..., H$), pipelines $\mathcal{P}_{h, k}$, ($k=1, 2, ..., G//H$), aim to optimize on datasets $\mathcal{D}_{h}$, and the best-performing pipelines $\mathcal{P}_{h, (1)}$ are selected by the ascending order of evaluation losses. The components of ensemble model $\Sigma_{H}$ can then be defined as $\mathcal{P}_{h}=\mathbf{P}_{h, (1)}$ and utilized to generate dataset $\mathcal{D}_{h+1}$ for next step. Following the residual learning framework, the boosting ensemble utilized in our work can be summarized as Algorithm \ref{alg:boosting}. The algorithm closely resembles Algorithm \ref{alg:bagging}, with the distinction of modifying its learning data set using residuals on response variables instead of employing feature subsets on feature matrices, as seen in the bagging ensemble. 
Similar to the bagging ensemble approach, the optimization of the residual dataset $\mathcal{D}_{h}$ can be fully parallelized. This leads to more efficient optimization and prevents the occurrence of sub-optimal pipelines on individual residual datasets.

{\SetAlgoNoLine
\begin{algorithm}
\caption{The Boosting Ensemble}\label{alg:boosting}
\KwIn{Dataset $\mathcal{D}=(\mathcal{D}_{train}, \mathcal{D}_{valid})$; Search space $\mathcal{U}$; Time budget $T$; Evaluation budget $G$; Search algorithm $Samp$; Size of the ensemble model $H$}
\KwOut{Ensemble $\Sigma_{M}$}
\textbf{Initialization}: $\textbf{y}_{train, 0}=\textbf{y}_{train}$;
$\textbf{y}_{valid, 0}=\textbf{y}_{valid}$;
$\hat{\textbf{y}}_{train, 0}=0$;
$\hat{\textbf{y}}_{valid, 0}=0$;\\
\For{$h \gets 1$ to $H$}{
$k= 0$ \Comment*[r]{Round of evaluation}
$t^{re} = T//H$ \Comment*[r]{Remaining time budget}
$g^{re} = G//H$ \Comment*[r]{Remaining evaluation budget}
$\textbf{y}_{train, h} = \textbf{y}_{train, h-1} - \hat{\textbf{y}}_{train, h-1}$;
$\textbf{y}_{valid, h} = \textbf{y}_{valid, h-1} - \hat{\textbf{y}}_{valid, h-1}$;\\
$\mathcal{D}_{h}=((\textbf{X}_{train}, \textbf{y}_{train, h}),(\textbf{X}_{valid}, \textbf{y}_{valid, h}))$;\\
\While{$t^{re} > 0$ and $g^{re} > 0$}{
    $t^{start}=CurrentTime$;\\
    $(E^{(k)}, \lambda_{E}^{(k)}), (I^{(k)}, \lambda_{I}^{(k)}), (B^{(k)}, \lambda_{B}^{(k)}), (S^{(k)}, \lambda_{S}^{(k)}), (F^{(k)}, \lambda_{F}^{(k)}), (M^{(k)}, \lambda_{M}^{(k)})=Samp^{(k)}(\mathcal{U})$;\\
    $\mathcal{P}_{h, k}=M^{(k)}_{\lambda_{M}^{(k)}}\circ F^{(k)}_{\lambda_{F}^{(k)}}\circ S^{(k)}_{\lambda_{S}^{(k)}}\circ B^{(k)}_{\lambda_{B}^{(k)}}\circ I^{(k)}_{\lambda_{I}^{(k)}}\circ E^{(k)}_{\lambda_{E}^{(k)}}$;\\
    $L^{eval, (k)}=\mathcal{V}(\mathcal{L}, \mathcal{P}_{h, k}, \mathcal{D}_{h})$;\\
    $t^{end}=CurrentTime$;\\
    $k = k + 1$;\\
    $t^{re} = t^{re} - (t^{end} - t^{start})$;\\
    $g^{re} = g^{re} - 1$;\\
}
$\{\mathcal{P}_{h, (k)}\}=sort(\{\mathcal{P}_{h, k}\})$;\\
$\hat{\textbf{y}}_{train, h}=\mathcal{P}_{h, (1)}(\textbf{X}_{train})$;
$\hat{\textbf{y}}_{valid, h}=\mathcal{P}_{h, (1)}(\textbf{X}_{valid})$;\\
}
$\Sigma_{H}=\Sigma_{H}(\mathcal{P}_{1, (1)}, \mathcal{P}_{2, (1)}, \ldots, \mathcal{P}_{H, (1)})$;\\
\Return{$\Sigma_{H}$};\\
\end{algorithm}}

\textbf{Voting Mechanism in Ensemble}: The voting mechanism, a critical component in the ensemble, determines how the predictions from individual candidate pipelines can be aggregated to produce the final prediction for the ensemble model. 

For the boosting ensemble, given their unique training diagram of residual learning, the voting mechanism is typically the summation of pipelines: 
$$
\hat{\textbf{y}}=\Sigma_{H}(\textbf{X})=\sum_{h=1}^{H}\mathcal{P}_{h}(\textbf{X})
$$

For stacking and bagging ensembles, the voting mechanisms must be tailored to the nature of the regression or classification tasks, but are interchangeable between stacking and bagging structures.

In regression tasks, the voting mechanisms often involve aggregation statistics. Commonly used statistics include mean, median, and maximum, calculated by performing the corresponding operation on all $H$ predictions by individual pipelines. For example, in the mean voting mechanism, the corresponding predictions of the ensemble models can be expressed as
$$
\hat{\textbf{y}}=\Sigma_{H}(\textbf{X})=\dfrac{1}{H}\sum_{h=1}^{H}\mathcal{P}_{h}(\textbf{X})
$$

In classification tasks, the voting can be either hard or soft \citep{polikar2006ensemble}, which differs by aggregating the class-level predictions or probabilities of each class. The soft voting aggregates the prediction probabilities generated by individual pipelines. For a $O$-category classification problem, the prediction probability of pipeline $h$, $\mathcal{P}_{h}$, can be represented as a probability vector
$$
\mathcal{P}_{h}^{prob}(\textbf{X})=(p_{h, 1}, p_{h, 2}, \ldots, p_{h, O})
$$
where $p_{h, o}\in [0, 1]$, ($o=1,2,\ldots,O$), denotes the probability of being categorised as class $o$ and $\sum_{o=1}^{O}p_{h, o}=1$. The ensemble model prediction in soft voting is the class with the highest aggregated probability:
$$
\hat{\textbf{y}}=\mathrm{argmax}\sum_{h=1}^{H}\mathcal{P}_{h}^{prob}(\textbf{X})=\mathrm{argmax}(\sum_{h=1}^{H}p_{h, 1}, \sum_{h=1}^{H}p_{h, 2}, \ldots, \sum_{h=1}^{H}p_{h, O})
$$
The hard voting, however, aggregates the prediction classes instead of the prediction probabilities. The prediction classes of individual pipeline $h$ can be expressed as
$$
\mathcal{P}_{h}(\textbf{X})=\mathrm{argmax}(p_{h, 1}, p_{h, 2}, \ldots, p_{h, O})
$$
indicating that the predicted class is the most probable class based on the probability. If class $o$ is the predicted class, the prediction can further be re-written as a unit vector 
\[
\begin{matrix}
\mathcal{P}_{h}(\textbf{X}) = o = 
    &(0,\ldots,0,&1,&0,\ldots,0)\\
    &&\uparrow&     \\
    &&o&            \\
\end{matrix}
\]
The aggregation of predictions in hard voting can be considered as the majority vote among all $H$ pipelines, with the probabilities reduced to indicator functions/unit vector representations. The eventual predictions of the ensemble model can be calculated as
\begin{align*}
\hat{\textbf{y}}&=\mathrm{argmax}(\sum_{h=1}^{H}\mathbbm{1}_{\{\mathcal{P}_{h}(\textbf{X})=1\}}, \sum_{h=1}^{H}\mathbbm{1}_{\{\mathcal{P}_{h}(\textbf{X})=2\}},\ldots,\sum_{h=1}^{H}\mathbbm{1}_{\{\mathcal{P}_{h}(\textbf{X})=O\}}) \\
&=\mathrm{argmax}\sum_{h=1}^{H}\mathcal{P}_{h}(\textbf{X})\\
\end{align*}
where the first row utilizes the numerical representation of $\mathcal{P}_{h}(\textbf{X})$ and second row corresponds to the unit vector representation. 

With the training diagram and the voting mechanism, multiple pipelines can be coordinated to address the imbalance problems that individual pipelines alone might struggle with, as demonstrated in various empirical studies \citet{polikar2006ensemble}, \citet{galar2011review}, \citet{dong2020survey}.

\clearpage

\section{Experiment code and description}\label{appendix_sec:code}

In the following, we present the code necessary to run the experiments demonstrated in Section \ref{sec:exp} and provide brief descriptions for each code block. The scripts for all experiments are available in our GitHub repository.\footnote{\url{https://github.com/PanyiDong/InsurAutoML/tree/master/experiments}}

\subsection{French Motor Third-Part Liability}\label{appendix_subsec:freMTPL2freq}

\begin{lstlisting}[language=Python, caption=French Motor Third-Part Liability Experiment Code, label=lst:freMTPL2freq]
import InsurAutoML
from InsurAutoML import load_data, AutoTabularRegressor
import numpy as np
import pandas as pd
from sklearn.metrics import mean_poisson_deviance

seed = 42
n_trials = 64
N_ESTIMATORS = 5
TIMEOUT = (n_trials / 4) * 450

InsurAutoML.set_seed(seed)

# load data
database = load_data(data_type = ".csv").load(path = "")
database_names = [*database]

# define response/features
response = "ClaimNb"
features = np.sort(list(
    set(database["freMTPL2freq"].columns) - set(["IDpol", "ClaimNb"])
))

# read train index & get test index
# python dataframe index starts from 0, but R starts from 1
train_index = np.sort(pd.read_csv("train_index.csv").values.flatten()) - 1
test_index = np.sort(
    list(set(database["freMTPL2freq"].index) - set(train_index))
)
# train/test split
train_X, test_X, train_y, test_y = (
    database["freMTPL2freq"].loc[train_index, features], database["freMTPL2freq"].loc[test_index, features], 
    database["freMTPL2freq"].loc[train_index, response], database["freMTPL2freq"].loc[test_index, response],
)


# fit AutoML model
mol = AutoTabularRegressor(
    model_name = "freMTPL2freq_{}".format(n_trials),
    n_estimators = N_ESTIMATORS,
    max_evals = n_trials,
    timeout = TIMEOUT,
    validation=False,
    search_algo="HyperOpt",
    objective= mean_poisson_deviance,
    cpu_threads = 12,
    balancing = ["SimpleRandomOverSampling", "SimpleRandomUnderSampling"],
    seed = seed,    
)
mol.fit(train_X, train_y)


train_pred = mol.predict(train_X)
test_pred = mol.predict(test_X)

mean_poisson_deviance(train_y, train_pred), mean_poisson_deviance(test_y, test_pred)
\end{lstlisting}

\begin{lstlisting}[language=R, caption=Generation of train set index, label=lst:train_idx]
RNGversion("3.5.0")
set.seed (100)
ll <- sample (c (1: nrow ( freMTPL2freq )) , round (0.9* nrow ( freMTPL2freq )) , replace = FALSE )
write.csv(ll, "train_index.csv") # the train_index.csv generated in R is utilized in AutoML train/test split
\end{lstlisting}

Listing \ref{lst:freMTPL2freq} presents the code used to run the French Motor Third-Part Liability experiment. The first 12 lines (Lines 1-12) are dedicated to preparing for the experiment. Specifically, Line 5 specifies the evaluation metric, which is the mean Poisson deviance. Lines 7-10 define key parameters such as the experiment's random seed, evaluation budget, the number of candidates used to construct the ensemble model, and the time budget. Lines 15-16 handle the data loading process, while Lines 19-20 define the response variable and the set of features. The train/test split process is executed between Lines 26-34, where the split is determined based on the train set index generated by Listing \ref{lst:train_idx}, following the procedure outlined by \citet{noll2020case}. Once the training is complete, predictions are made on both the train and test sets, with the train/test mean Poisson deviance being reported as indicated in Lines 53-56.

In the training setup, we assign the experiment name based on the evaluation budget, as indicated in Line 39. For instance, if the evaluation budget is set to 512, the experiment will be named \textit{freMTPL2freq\_512}. Upon completion of the training, a folder with the same name is created to store all the experiment results, along with a file of the same name containing the final ensemble model. The experiment configuration, including the components of the ensemble model, the evaluation budget, the time budget, the cross-validation methodology, the search algorithm, the evaluation metric, the number of parallel computing threads, the balancing algorithms, and the random seed, are defined in Lines 40-48. Specifically, we employ the HyperOpt search algorithm as described by \citet{Bergstra2013}. Due to computational constraints, we limit the balancing algorithms to simple random over-sampling and under-sampling techniques.

\subsection{Wisconsin Local Government Property Insurance Fund}\label{appendix_subsec:LGPIF}

\begin{lstlisting}[language=Python, caption=Wisconsin Local Government Property Insurance Fund Experiment Code, label=lst:LGPIF]
import InsurAutoML
from InsurAutoML import load_data, AutoTabularRegressor
import numpy as np
from sklearn.metrics import r2_score

seed = 42
n_trials = 64
N_ESTIMATORS = 5
TIMEOUT = (n_trials / 4) * 450

InsurAutoML.set_seed(seed)

# load data
database = load_data(data_type = ".rdata").load(path = "")
database_names = [*database]

# define response/features
response = ["yAvgBC"]
features = [
    'TypeCity', 'TypeCounty', 'TypeMisc', 'TypeSchool', 'TypeTown', 'TypeVillage', 'IsRC', 'CoverageBC', 'lnDeductBC', 
    'NoClaimCreditBC', 'CoverageIM', 'lnDeductIM', 'NoClaimCreditIM', 'CoveragePN', 'NoClaimCreditPN', 'CoveragePO', 
    'NoClaimCreditPO','CoverageCN', 'NoClaimCreditCN', 'CoverageCO', 'NoClaimCreditCO'
]
# log transform of response
database["data"][response] = np.log(database["data"][response] + 1)
database["dataout"][response] = np.log(database["dataout"][response] + 1)
# log transform of coverage feateres
database["data"][["CoverageBC", "CoverageIM", "CoveragePN", "CoveragePO", "CoverageCN", "CoverageCO"]] = np.log(
    database["data"][["CoverageBC", "CoverageIM", "CoveragePN", "CoveragePO", "CoverageCN", "CoverageCO"]] + 1
)
database["dataout"][["CoverageBC", "CoverageIM", "CoveragePN", "CoveragePO", "CoverageCN", "CoverageCO"]] = np.log(
    database["dataout"][["CoverageBC", "CoverageIM", "CoveragePN", "CoveragePO", "CoverageCN", "CoverageCO"]] + 1
)

train_X, train_y = database["data"][features], database["data"][response]
test_X, test_y = database["dataout"][features], database["dataout"][response]

# fit AutoML model
mol = AutoTabularRegressor(
    model_name = "LGPIF_{}".format(n_trials),
    n_estimators = N_ESTIMATORS,    
    max_evals = n_trials,
    timeout = TIMEOUT,
    validation="KFold",
    valid_size=0.2,
    search_algo="HyperOpt",
    objective= "R2",
    cpu_threads = 12,
    seed = seed,    
)
mol.fit(train_X, train_y)

train_pred = mol.predict(train_X)
test_pred = mol.predict(test_X)
r2_score(train_y, train_pred), r2_score(test_y, test_pred)
\end{lstlisting}

Listing \ref{lst:LGPIF} is the code used to run the Wisconsin Local Government Property Insurance Fund experiments. The structure of this experiment follows the same setup as previously described in the French Motor Third-Party Liability experiment, including environment configuration, data loading, definition of features and response variables, AutoML fitting, and prediction generation. Unlike the French Motor Third-Part Liability experiment, the LGPIF dataset includes both in-sample and out-of-sample data, eliminating the need for a train/test split process. After reading the two Rdata files, we replicate the preprocessing steps outlined by \citet{quan2018predictive} by applying logarithmic transformations to the response variable and selected features.

In the training setup, we employ five-fold cross-validation, using \textit{validation} as \textit{KFold} with \textit{valid\_size} set to $0.2$. Given the smaller data size, the restrictions on balancing algorithms applied in the previous experiment are not enforced in this experiment. Additionally, the evaluation metric selected for this experiment is the $R^{2}$ score.

\subsection{Australian Automobile Insurance}\label{appendix_subsec:ausprivauto}

\begin{lstlisting}[language=Python, caption=Australian Automobile Insurance Experiment Code, label=lst:ausprivauto]
import pandas as pd
import InsurAutoML
from InsurAutoML import load_data, AutoTabular
from InsurAutoML.utils import train_test_split

seed = 42
n_trials = 128
N_ESTIMATORS = 4
TIMEOUT = (n_trials / 4) * 450

InsurAutoML.set_seed(seed)

# load data
database = load_data(data_type = ".csv").load(path = "")
database_names = [*database]

# define response/features
response = "ClaimOcc"
features = list(
    set(database["ausprivauto"].columns) - set(["ClaimOcc", "ClaimNb", "ClaimAmount"])
)
features.sort()

# train/test split
train_X, test_X, train_y, test_y = train_test_split(
    database['ausprivauto'][features], database['ausprivauto'][[response]], test_perc = 0.1, seed = seed
)
pd.DataFrame(train_X.index.sort_values()).to_csv("train_index.csv", index=False)

# fit AutoML model
mol = AutoTabular(
    model_name="ausprivauto_occ_{}".format(n_trials),
    max_evals=n_trials,
    n_estimators=N_ESTIMATORS,
    timeout=TIMEOUT,
    validation="KFold",
    valid_size=0.25,
    search_algo="Optuna",
    objective="AUC",
    cpu_threads=12,
    seed=seed,
)
mol.fit(train_X, train_y)

from sklearn.metrics import roc_auc_score

y_train_pred = mol.predict_proba(train_X)
y_test_pred = mol.predict_proba(test_X)
roc_auc_score(train_y.values, y_train_pred["class_1"].values), roc_auc_score(test_y.values, y_test_pred["class_1"].values)
\end{lstlisting}

The code for running the Australian Automobile Insurance experiment is detailed in Listing \ref{lst:ausprivauto}. As this is a classification task, we utilize \textit{AutoTabular} with automatic task type selection. Initially, we perform a 90/10 train/test split for the first experiment and reuse the generated train set index for subsequent experiments. The experimental setup involves four-fold cross-validation, employing the Optuna \citep{Akiba2019} search algorithm and using AUC as the evaluation metric.

\end{appendices}

\end{document}